\documentclass[10pt,twocolumn,letterpaper]{article}

\usepackage[pagenumbers]{cvfconf}

\definecolor{ieeeblue}{rgb}{0.21,0.49,0.74}

\usepackage[pagebackref,breaklinks,colorlinks,allcolors=ieeeblue]{hyperref}

\usepackage[linesnumbered,ruled,vlined]{algorithm2e}

\usepackage{multirow}
\usepackage{makecell}
\usepackage{url}
\usepackage{amsmath,amsthm}
\newtheorem{theorem}{Theorem}[section] 
\newtheorem{definition}[theorem]{Definition}

\newtheorem{proposition}[theorem]{Proposition}

\usepackage[capitalize]{cleveref}
\crefname{section}{Sec.}{Secs.}
\Crefname{section}{Section}{Sections}
\Crefname{table}{Table}{Tables}
\crefname{table}{Tab.}{Tabs.}

\usepackage{bbding}

\makeatletter
\def\blfootnote{\xdef\@thefnmark{}\@footnotetext}
\makeatother

\def\paperID{} 
\def\confName{}
\def\confYear{}

\title{Lipschitz Constant Meets Condition Number: \\
Learning Robust and Compact Deep Neural Networks}

\author{Yangqi Feng$^{1,*}$, Shing-Ho J. Lin$^{2,*}$, Baoyuan Gao$^3$, Xian Wei$^{1,\dagger}$\\
$^1$Software Engineering Institute, East China Normal University\\
$^2$School of Artificial Intelligence, University of Chinese Academy of Sciences \\
$^3$Tianjin University 
}

\begin{document}
\maketitle

\blfootnote{$^*$Equal Contribution. $^\dagger$Corresponding Author.}

\begin{abstract}
Recent research has revealed that high compression of Deep Neural  Networks (DNNs), e.g., massive pruning of the weight matrix of a DNN, leads to a severe drop in accuracy and susceptibility to adversarial attacks. 
Integration of network pruning into an adversarial training framework has been proposed to promote adversarial robustness. 
It has been observed that a highly pruned weight matrix tends to be ill-conditioned, i.e., increasing the condition number of the weight matrix. 
This phenomenon aggravates the vulnerability of a DNN to input noise. 
Although a highly pruned weight matrix is considered to be able to lower the upper bound of the local Lipschitz constant to tolerate large distortion, the ill-conditionedness of such a weight matrix results in a non-robust DNN model. 
To overcome this challenge, 
this work develops novel joint constraints to adjust the weight distribution of networks, namely, the Transformed Sparse Constraint joint with Condition Number Constraint (TSCNC), which 
copes with smoothing distribution and differentiable constraint functions to reduce condition number and thus avoid the ill-conditionedness of weight matrices. 
Furthermore, our theoretical analyses unveil the relevance between the condition number and the local Lipschitz constant of the weight matrix, namely, the sharply increasing condition number becomes the dominant factor that restricts the robustness of over-sparsified models. 
Extensive experiments are conducted on several public datasets, and the results show that the proposed constraints significantly improve the robustness of a DNN with high pruning rates.
\end{abstract}

\section{Introduction}
\begin{figure}
    \centering
    \includegraphics[width=0.48\textwidth]{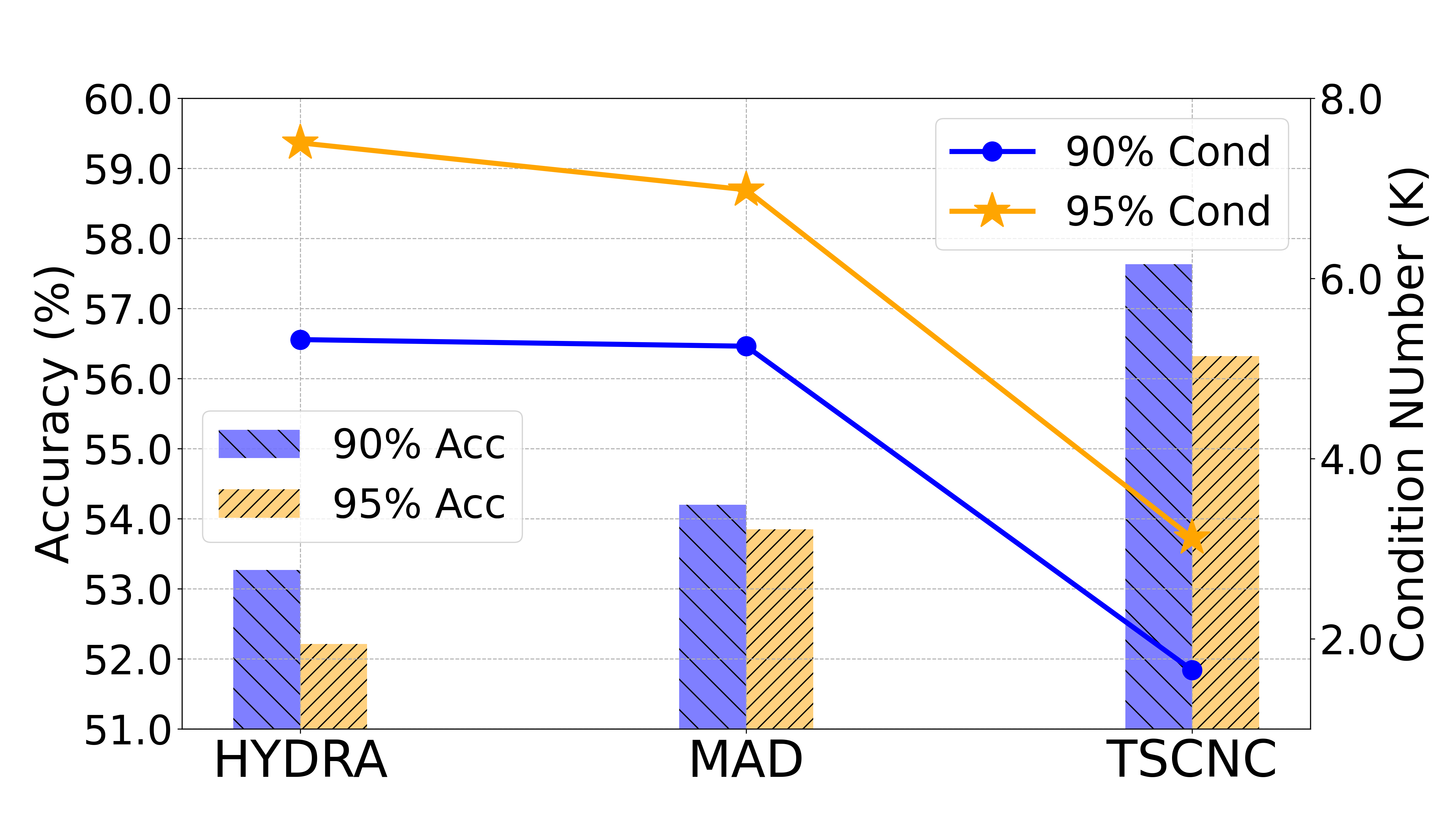}
    \caption{Comparison of condition number and robust accuracy of three methods at different pruning rates (90\% v.s.
95\%). The histogram 'Acc' represents the value of the robust accuracy of different methods at the same pruning rate, while the line graph 'Cond' represents the condition number of the model.}
    \label{comparsion_acc}
\end{figure}
\begin{figure*}[h]
	\centering
	\begin{subfigure}{0.325\linewidth}
		\centering
		\includegraphics[width=1.0\linewidth]{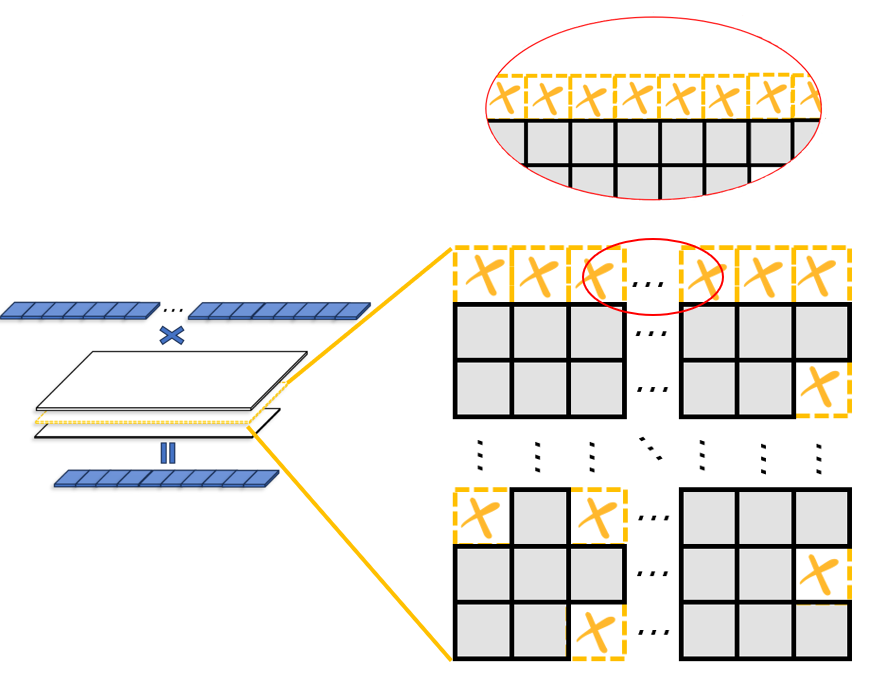}
		\caption{Ill-conditioned weight space}
		\label{ill-condition}
	\end{subfigure}
	\centering
	\begin{subfigure}{0.325\linewidth}
		\centering
		\includegraphics[width=0.9\linewidth]{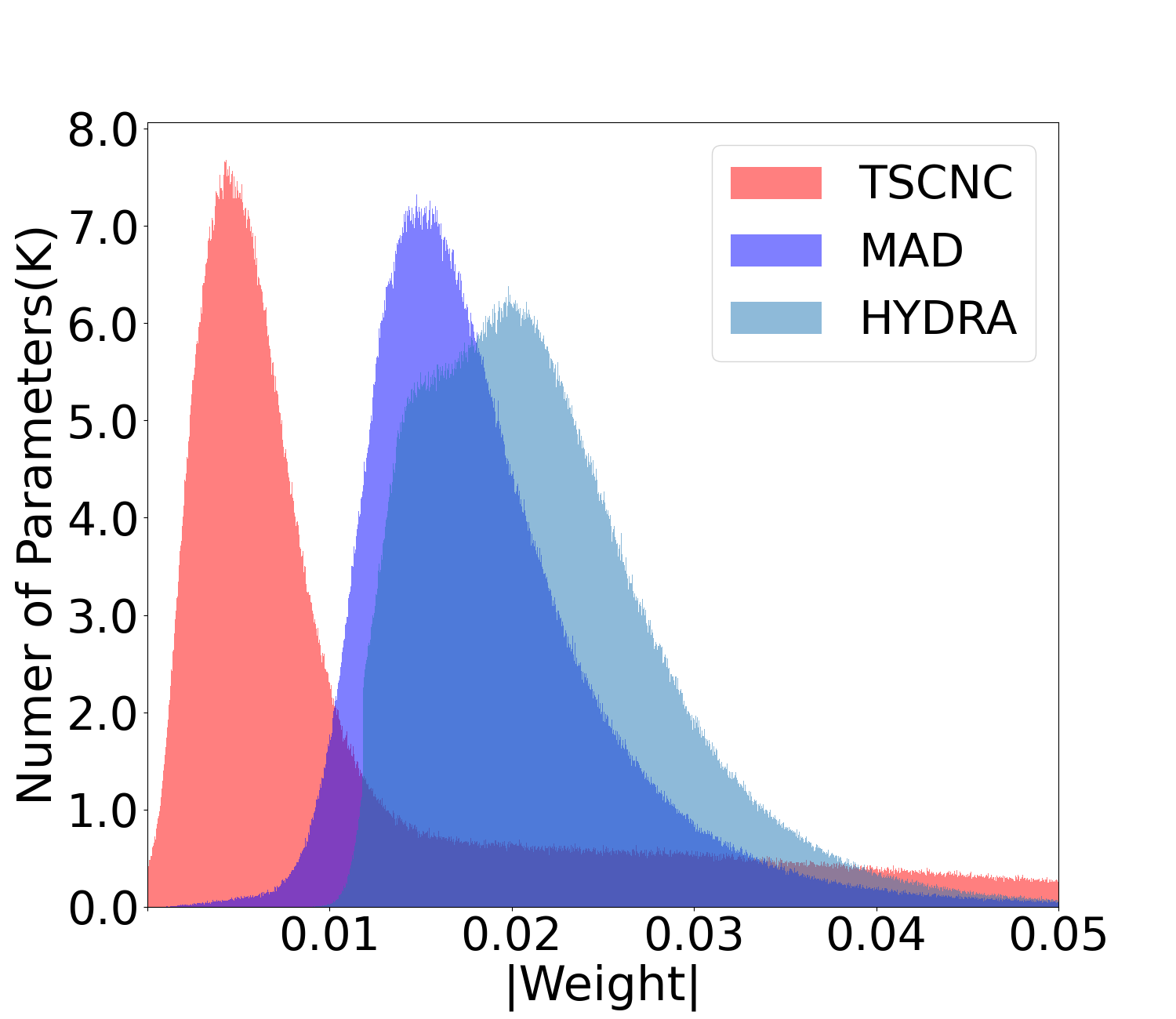}
		\caption{90\% sparsity}
		\label{90sparsity}
	\end{subfigure}
	\centering
	\begin{subfigure}{0.325\linewidth}
		\centering
		\includegraphics[width=0.9\linewidth]{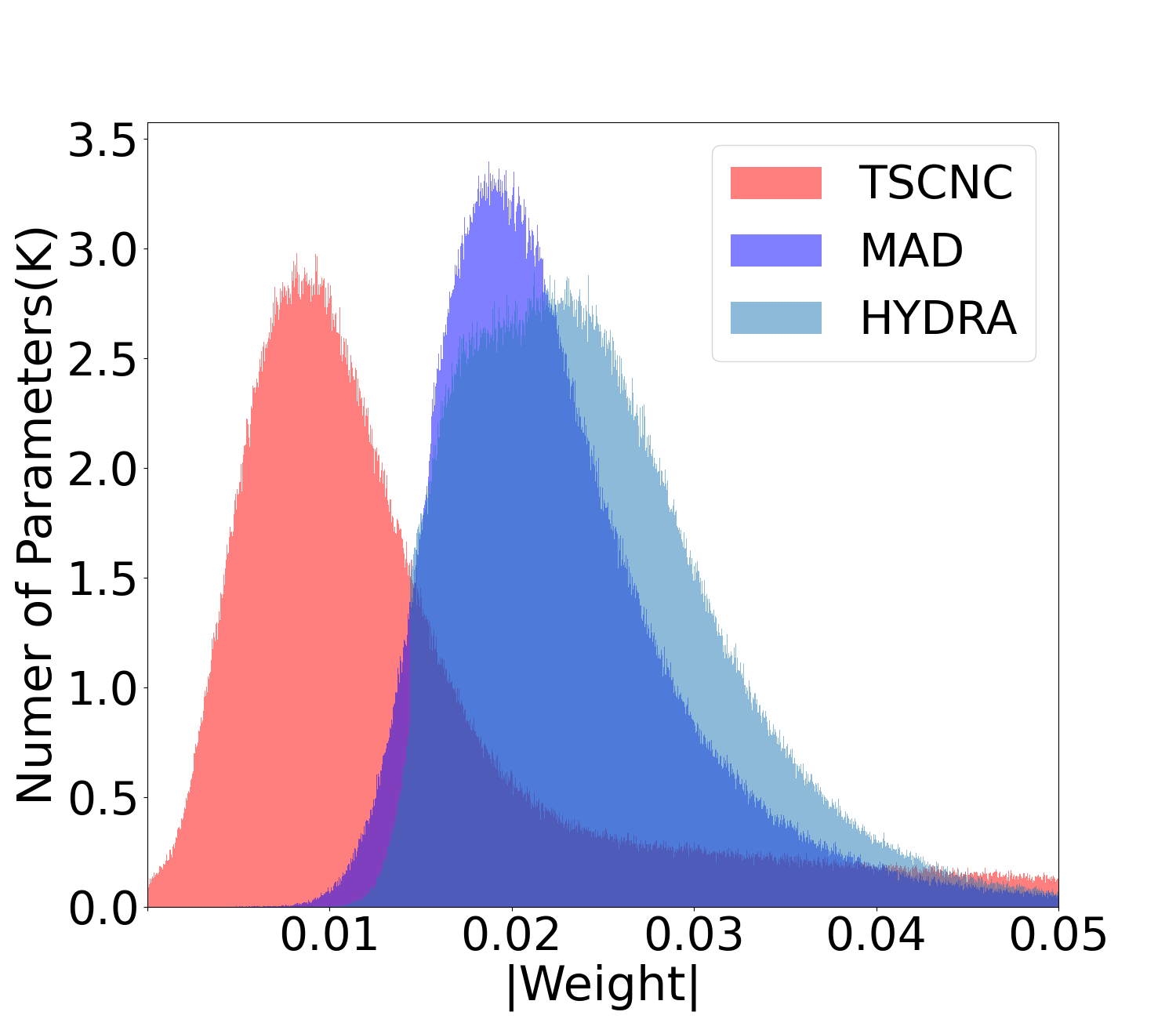}
		\caption{95\% sparsity}
		\label{95sparsity}
	\end{subfigure}
	\caption{a) The ill-conditioned weight space in the full connection layer; b) The parameter distribution of the pruned model on $90\%$ sparsity, the value distribution; c) The parameter distribution of the pruned model on $95\%$ sparsity. See per-layer pruning ratio in \textit{Supplementary Materials}.}
	\label{da_chutian}
\end{figure*}

With an increasing depth of deep neural networks (DNNs), the size of a DNN has exploded to encompass millions or even billions of parameters, introducing significant challenges for resource-constrained devices, marked by substantial computational and storage burdens, stemming from highly redundant feature representations and over-parameterization.
It is worth noticing that even in a sparse network configuration, comprising merely $5\%$ of the total parameters has demonstrated comparable performance to that of fully parameterized networks \cite{denil2013predicting} in the best case. 
Furthermore, previous works \cite{guo2016dynamic,ullrich2017soft,han2015learning,anwar2017structured} discover that well-trained DNNs exhibit the potential for pruning more than $90\%$ of their connections without incurring any discernible loss in accuracy. 
These results suggest that the pruned redundant neurons are of 
minor of the least contribution to the model's performance. 
However, the aforementioned sparse models, i.e., the pruning models with only $10\%$ or fewer neuron connections left, are extremely susceptible to carefully crafted perturbations, i.e., adversarial examples \cite{wei2022learning}. 
To address this concern, PGD \cite{madry2017towards} suggests that network capacity is crucial to robustness, as larger network capacity implies more complex adversarial decision boundaries. 
Increasing the network capacity may provide a better trade-off between standard accuracy and its adversarial robustness \cite{tsipras2018robustness}. 
In contrast, it is notable that an appropriately higher weight sparsity \cite{guo2018sparse,xiao2018training} implies better robustness on naturally trained models. 
It seems that the two viewpoints above conflict with each other, but after providing a comprehensive analysis from both theoretical and empirical perspectives, we find that these discordant viewpoints are essentially consistent.

To obtain robust and reliable networks, Min-Max robust optimization-based Adversarial Training (AT) \cite{madry2017towards,tsipras2018robustness} provides a popular way against first-order adversarial attacks that rely on gradients of the loss function w.r.t. the input. 
The studies \cite{ye2019adversarial,sehwag2020hydra} incorporate pruning strategies and AT techniques into a unified optimization framework, bolstering sparse model robustness. 
The framework mentioned above can indeed alleviate the decreasing robustness due to sparsity. 
However, it is unclear which connections in the network are critical for preserving model performance, and the robust accuracy sharply drops along with the increasing model sparsity. 
To elucidate these aspects, we designed a comprehensive set of experimental and theoretical analyses.

In this work, we first study the relationship between the sparsity and the adversarial robustness of DNNs and develop constraints to alleviate the contradictions between model pruning and robustness. We then theoretically prove that the proposed sparse constraint increases the sparsity of a DNN by limiting the Lipschitz constant. 
However, over-sparsified models undermine robustness due to dramatically increasing condition number of the weight matrix, yielding an ill-conditioned weight space \cite{wei2022learning}. 
We observed that the condition number of the weight matrices replaces the Lipschitz constant when the sparsity rises, even though the sparsity constraint has limited the Lipschitz constant. 
Empirical evaluation on CIFAR10 \cite{krizhevsky2009learning}, CIFAR100 \cite{krizhevsky2009learning} and Tiny-Imagenet \cite{le2015tiny} validates our approach's performance and effectiveness, which coincides with our motivations and theory. 

The main contributions of this paper are summarized as
follows:
\begin{itemize}
    \item To the best of our knowledge, we are the first to theoretically prove the relationship among the sparsity of the model, the local Lipschitz constant, and the condition number of weight matrices of a DNN.
    \item Based on the theoretical findings, we developed novel differentiable constraints for training a sparse and robust network, namely, Transformed Sparse Constraint joint with Condition Number Constraint (TSCNC), to resolve the conflict between the model's sparsity and robustness.
    \item In experiments, we achieve state-of-the-art performance in mainstream (VGG, ResNet, and WideResNet) architectures on various datasets. Besides, it demonstrates strong migratability and the visualization of condition number on the last layers indicates that our method improves the robustness against adversarial samples.
\end{itemize}

\section{Related Work}
In this section, we briefly review sparse optimization-based
methods, network pruning, and adversarial training.

\subsection{Sparse Constraint} 
Finding sparse networks via pruning useless neuronal connections is a popular solution in the sparsity of DNNs \cite{gui2019model,madaan2020adversarial,chen2022linearity,PAMI24Pruning,PAMI24Pruning2}.
The pretraining-pruning fine-tuning three-step pruning strategy \cite{han2015deep} is a widely used training scheme.
However, pre-training a dense network is time-consuming, thus it can train a sparse network directly from scratch through ${\ell _{p}}$ norm regularizer $(p\in \{0,1\})$. Consider an empirical risk minimization of the general objective function with a sparse ${\ell _{p}}$ constraint term, as shown in~\cref{eq:1}:
\begin{equation}
  \mathcal{L}(\theta) = \arg {\min\limits_{\theta}}\frac{1}{N}\left(\sum_{i=1}^{N}\mathcal{L}\left(h\left(x_{i};\theta\right),y_{i}\right)\right)+\lambda\left\| \theta \right\|_{p},
  \label{eq:1}
\end{equation}
where $\left\|\theta  \right\|_{p}$ denotes the ${\ell _{p}}$ norm regularization of the parameters $\theta $ and $\lambda$ is a weighing factor for regularization to control the sparsity of  $\theta $. $\mathcal{L}\left( \cdot \right)$ is the loss function and ($x_{i}$,$y_{i}$) denotes input-output pairs composed of dataset of size $N$.

Among them, ${\ell _{1}}$ norm regularization (Lasso) ~\cite{kukreja2006least} forces the value of the weight matrix to be close to zero with the advantages of feature selection and interpretability, has been widely used to train a sparse DNNs. In contrast to ${\ell _{1}}$ regularizer, ${\ell _{0}}$ norm regularization term~\cite{louizos2017learning} imposes a stricter penalty of sparsity constraint on the weight matrix and ${\ell _{0}}$ norm causes no shrinkage on the actual values of the weight matrix. However, minimizing ${\ell _{0}}$ norm regularization term is usually numerically intractable and NP-hard~\cite{natarajan1995sparse}.

\subsection{Adversarial Training} 
To improve the robustness of a network against well-designed adversarial examples,
AT~\cite{denil2013predicting} is one of the primarily effective approaches in practice among previous work~\cite{biggio2013evasion,goodfellow2014explaining,carlini2017towards}. AT uses a natural saddle point formulation as the objective function to optimize the Min-max empirical risk problem:
\begin{equation}
   \min_{\theta}{ \mathbb{E}_{(\mathcal{X},\mathcal{Y})\sim \mathcal{D}}\left[ \underset{{\left\| \delta \right\|}_{p}\leq \varepsilon }{\max}\mathcal{L}\left( \theta ,\mathcal{X}+\delta,\mathcal{Y} \right)\right]},
   \label{eq:2}
\end{equation}
where $(\mathcal{X},\mathcal{Y})$ is the input-output pairs of datasets submit to distribution $\mathcal{D}$, $\mathcal{L}(\cdot)$ is the loss function and $\theta$ represents the set of weight parameters, $\delta$ is additive adversarial perturbation within $\varepsilon \subseteq \mathbb{R}^{d}$.
The inner maximization problem aims to find the adversarial perturbation to input samples that achieves a high loss by iterative adversarial attacks, such as  Projected Gradient Descent (PGD) based attacks~\cite{carlini2017towards}, the outer minimization problem is to obtain model parameters so that the “adversarial loss” given by the inner attack problem is minimized.

\subsection{Learning Sparse and Robust Models} 
Recent works are devoted to constructing simultaneously sparse and robust models by using sparse settings and adversarial training frameworks. 
\cite{sehwag2020hydra} proposed to make the pruning process aware of the robust training objective and remove slight impact connects according to AT loss. 
\cite{lee2022masking} proposed to train sparse and robust DNNs through second-order information of adversarial loss.
\cite{wei2022learning} proposed to combine the tensor product with the differentiable constraints for promoting sparsity and improving robustness.
\cite{cosentino2019search} proposed an inherent trade-off between sparsity and robustness, finding sparse and robust DNNs through empirical evaluation and an analytical test of the Lottery Ticket Hypothesis.

However, the aforementioned pruning strategies backward propagation of AT loss, which tend to model parameters not towards exactly zero, thereby, leading to pruning parameters with small coefficients, while these parameters may contribute to the robustness of DNNs. Sparse methods can still improve robustness through limiting Lipshitz constant \cite{guo2018sparse}. Inevitably, their robustness drops dramatically at high sparsity. An attractive problem is that higher sparse models are more vulnerable to adversarial samples.
Thus, it is necessary to demonstrate the relevance between the sparsity and robustness of networks, and the intrinsic factors that affect the balance of sparse networks to adversarial samples. 

\section{Methodology}
In this section, we present the details of the proposed TSCNC. 
We first introduce the relevance between the higher sparsity and the robustness, then propose a novel sparse constraint that is consistent with our theoretical analysis, preventing the dramatic deterioration of robust accuracy in extremely sparse situations.

\subsection{Sparsity and Robustness}

We focus on DNNs with ReLU activation functions for classification tasks as an example to study the relevance between sparsity and robustness, the same applies to the complex structure of some modern DNNs as well as most pooling and activation functions being predefined with fixed policies. 
Let us consider a simple multi-layer perceptron $g\left( \cdot \right)$ with $L$ hidden layers and each hidden layer is parameterized by a weight matrix $\mathbf{W}_{l}\in \mathbb{R}^{n_{l-1}\times n_{l}}$ and $w_{k}\in \mathbf{W}_{L+1}\left[:,k \right]$, in which $l\in[1:L+1]$. 
For input samples $x_{i}$, we have
\begin{equation}
\begin{aligned}
g_{k}\left(\mathbf{x}_{i}\right) = 
\mathbf{w}_{k}^{\top} \sigma\left(
\mathbf{W}_{L-1}^{\top} \sigma\left(\cdots \sigma\left(\mathbf{W}_{1}^{\top}\mathbf{x}_{i}\right)\right)
\right)
\label{eq:3}
\end{aligned}
\end{equation}
where $\mathbf{w}_{k}$ is the $k$-th column of $\mathbf{W}_L$, $g_{k}\left( \cdot \right)$ denotes the prediction scores of input samples for class $k$ and $\sigma$ is the activation function. Here, we ignore the bias term $\mathbf{b}$ for clarity. Notice that $\mathbf{W}^{\top}\mathbf{x} + \mathbf{b}$ can be simply
rewritten as  $\widetilde{\mathbf{W}}^{\top}[\mathbf{x}; 1]$, where $\widetilde{\mathbf{W}} = [\mathbf{W}; \mathbf{b}]$. 
Let $\hat{y}=\arg \max_{k \in [1:c]} g_{k}\left(\mathbf{x}_{i}\right)$ denote the predicted output classification results. 
Assuming the classifier is Lipschitz continuous, let us denote by $L_{q,\mathbf{x}}^{k}$ the (optimal) local Lipschitz constant of the function $g_{\hat{y}}\left(\mathbf{x}\right)-g_{k}\left(\mathbf{x}\right)$ over a fixed $B_{p}\left(\mathbf{x},r\right)$, in which $B_{p}\left(\mathbf{x},r\right)$ is a close ball centered at $x$ with radius $r>0$ under $\ell_{p}$ norm. 
Based on previous work~\cite{guo2018sparse,hein2017formal}, the relation between the local Lipschitz constant and the weights is given as follows: 
\begin{proposition}[\cite{weng2018evaluatingl}]
\label{Prop:3.1}
Let $\hat{y}=\arg \max_{k \in [1:c]} g_{k}\left(\mathbf{x}\right)$, and $\frac{1}{p}+\frac{1}{q}=1$, then for any $ \Delta_{\mathbf{x}} \in B_{p}\left( \mathbf{0}, r\right) $, $p \in \mathbb{R}^+$ and a set of Lipschitz continuous functions $\{g_k:\mathbb{R}^n \to \mathbb{R} \} $, with:
\begin{equation}
\begin{aligned}
\left\| \Delta_{\mathbf{x}} \right\|_p \leq 
\min \left\{ \min_{k \neq \hat{y}}{ \dfrac{g_{\hat{y}}(\mathbf{x}) - g_k(\mathbf{x})}{L_{q,\mathbf{x}}^{k}}, r} \right\} := \gamma
\label{eq:4}
\end{aligned}
\end{equation}
it holds that $\hat{y}=\arg \max_{k \in [1:c]} g_{k}(\mathbf{x} + \Delta_{\mathbf{x}})$, which means the classification decision does not change on $B_p(\mathbf{x}, \gamma)$. Therein, $\gamma$ is an instance-specific lower bound that guarantees the robustness of multi-class classifiers.
\end{proposition}

\begin{theorem}[Sparsity and Robustness of nonlinear DNN~\cite{guo2018sparse}]
\label{Theorem:3.2}
Let the weight matrix be represented as $\mathbf{W}_j = \mathbf{W}^\prime_j \circ \mathbf{M}_j$, in which $\{\mathbf{M}_j\left[u,v\right]\}$ are independent non-zero Bernoulli $B(1, 1 - \alpha_j )$ random variables, for $j \in [1:l-1]$. Then for any $\mathbf{x}\in \mathbb{R}^n$ and $k \in [1:c]$, it satisfies the following:
\begin{equation}
\mathrm{E}_{M_{[1:d-1]}}(L_{i,\mathbf{x}}^{k}) \leq 
c_i \left(1 - \eta(\alpha_1 ,..., \alpha_{l-1};\mathbf{x}) \right)
\end{equation}
where $i=1,2$, and $\eta$ is monotonically increasing function w.r.t. each $\alpha_j$.
Here, $c_2 = \left\|\mathbf{W}_{\hat y} - \mathbf{W}_k \right\|_2 \prod_j \left\| \mathbf{W}^\prime_j \right\|_F$ and 
$c_1 = \left\|\mathbf{w}_{\hat{y}} - \mathbf{w}_{k} \right\|_1 \prod_j \left\| \mathbf{W}^\prime_j \right\|_{1,\infty}$ 
are two constants,
\end{theorem}

According to~\cref{Theorem:3.2}, for $q \in \{1, 2\} \; (i.e., p \in \{\infty, 2\})$, $L_{q,\mathbf{x}}^{k}$ is prone to get smaller if any weight matrix gets more sparse.
At the same time, if the distribution of weights is closer to zero, it suggests a smaller value of $\left\|\textbf{W}_{j}\right\|_{p}$.
It is worth noting that the Lipschitz constant is of great importance for evaluating the robustness of DNNs, and it is effective to regularize DNNs by minimizing $L_{q,\mathbf{x}}^{k}$, or equivalently $\left\|\bigtriangledown g_{\hat{y}}\left(\mathbf{x}\right) -\bigtriangledown g_{k}\left(\mathbf{x}\right)\right\|_q$ for differentiable continuous functions \cite{hein2017formal}.
A smaller Lipschitz constant implying larger $\gamma$ and stronger robustness represents a higher level of robustness as a larger distortion can be tolerated. An appropriately higher weight sparsity or smaller parameter distribution 
of weights leading to a high level of robustness.

However, when we continue to apply stronger sparse constraints or pruning strength to get a smaller value of $\left\|\mathbf{W}_{j}\right\|_{p}$, we observe that the DNNs cannot get more robust by limiting the local Lipschitz constant, even contrarily, leading to dramatic drops in robustness.
This is because unstructured sparsity may lead to a non-full rank of the weight matrix with a certain row or column of all zero elements, as shown in~\cref{ill-condition}. The condition number of a low-rank matrix is $\infty$, which means the matrix falls into ill-conditioned weight space.

\subsection{Lipschitz Constant and Condition Number}

\begin{definition}[$\ell_2$-norm condition number]
The $\ell_2$-norm condition number of a full-rank matrix $\mathbf{A}\in \mathbb{R}^{K\times I}$ is defined as: $\kappa(\mathbf{A}) = \frac{\sigma_{\max}(\mathbf{A})}{\sigma_{\min}(\mathbf{A})}$, where $\sigma_{\max}(\mathbf{A})$ and $\sigma_{\min}(\mathbf{A})$ are maximal and minimal singular values of $\mathbf{A}$, respectively.
\end{definition}

The condition number $\kappa(\mathbf{A})$ is commonly used to measure the sensitivity of a function in the event of how much a small perturbation of the input $\mathbf{x}$ can change the output $\mathbf{y}$. A matrix with the condition number being close to one is said to be ``well-conditioned'', on the other side a matrix with a large condition number is said to be ``ill-conditioned'', which causes the vanishing and exploding gradient problem \cite{sinha2018neural_EKDD}. Let us consider an input perturbation $\delta \mathbf{x}$ and output perturbation $\delta \mathbf{y}$, which satisfies $\mathbf{y}+\delta\mathbf{y}=\mathbf{W}(\mathbf{x}+\delta\mathbf{x})$. Therein, $\mathbf{W}=\{\mathbf{W}_l\}$ are weight matrix.
Then we have
\begin{equation}
    \frac{1}{\kappa(\mathbf{W})}
    \frac{\left\|\delta\mathbf{x}\right\|}{\left\|\mathbf{x}\right\|}
    \leq 
    \frac{\left\|\delta\mathbf{y}\right\|}{\left\|\mathbf{y}\right\|}
    \leq
    \kappa(\mathbf{W})\frac{\left\|\delta\mathbf{x}\right\|}{\left\|\mathbf{x}\right\|}
    \label{condition_number}
\end{equation}
From Eq.~\eqref{condition_number}, it can be concluded that $\kappa(\mathbf{W})$ restricts the range of variation of output perturbation $\delta\mathbf{y}$. That is, improving the condition number promotes the robustness of the network to the adversarial noise.

We then have the following theorem.
\begin{theorem}
The sparsity of DNNs satisfies the following relationship with local Lipschitz constant $L_{q,\mathbf{x}}^{k}$ and the condition number $\kappa\left( \mathbf{W}\right)$:
\begin{equation}
 \begin{aligned}
\frac{1}{2}\cdot \frac{L_{q,\mathbf{x}}^{k}}{\left\|\mathbf{W} \right\|} \leq \kappa\left(\mathbf{W}\right)
 \label{eq:Lipschitz_condition}
 \end{aligned}
\end{equation}
\begin{proof}[Proof (Sketch)]
By getting the RHS of the Eq. \eqref{condition_number} and changing the position of $\left\| \mathbf{x}\right\|$ and $\left\|\delta \mathbf{y}\right\|$, we have,
\begin{equation}
    \frac{\left\|\mathbf{x}\right\|}{\left\|\mathbf{y}\right\|} \leq {\kappa (\mathbf{W})} \frac {\left\| \delta \mathbf{x}\right\|}{\left\| \delta \mathbf{y}\right\|}
    \label{eq:12}
\end{equation}
Next substituting an equation $y=\mathbf{W}x$ into Eq.\eqref{eq:12}, then we can get,
\begin{equation}
    \label{eq:pattern}
    \frac{1}{\left\|\mathbf{W}\right\|} 
    \leq \kappa (\mathbf{W}) 
    \frac{\left\|\delta \mathbf{x}\right\|}{\left\| \delta \mathbf{y}\right\|}
\end{equation}
Without loss of generality, if we keep the relative error $\frac{\left\|\delta \mathbf{y}\right\|}{\left\|\delta\mathbf{x}\right\|}$ of the model at a constant value, the lower bound of condition number $\kappa (\mathbf{W})$ will become large when we increase the sparsity of the model, i.e.,  the small value of $\left\| \mathbf{W}\right\|$. In contrast, we can limit the condition number of the weight matrix to improve the robustness through differentiable constraints at high sparsity. Thus we need to restrain the relative error $\frac{\left\|\delta \mathbf{y}\right\|}{\left\|\delta\mathbf{x}\right\|}$ to a small value.
Let us denote $h(\cdot ):=g_{\hat{y}}(\cdot )-g_{k}(\cdot )$, according to the definition of Lipschitz's constant, the $L_{q,\mathbf{x}}^{k}$ can be re-expressed as follows:
\begin{equation}
    L_{q,\mathbf{x}}^{k} = 
    \underset{\|\delta\mathbf{x}\|\leq {r}}{\rm sup}
    \frac{\|h(\mathbf{x} + \delta \mathbf{x}) - h(\mathbf{x})\|}{\|\delta \mathbf{x}\|}
    \label{eq:lipsizi_10}
\end{equation}
Moreover, the upper bound of $\|h_{k}(\mathbf{x} + \delta \mathbf{x}) - h_{k}(\mathbf{x})\|$ can be obtained, which is equivalent to the maximum,
\begin{equation}
\begin{aligned}
    & \left\| h_{k}(\mathbf{x} + \delta \mathbf{x}) - h_{k}(\mathbf{x})\right\| 
\\
   =& \left\| 
   \left[g_{\hat{y}}(\mathbf{x} + \delta \mathbf{x}) - g_{k}(\mathbf{x} + \delta \mathbf{x})\right] - 
   \left[g_{\hat{y}}(\mathbf{x}) - g_{k}(\mathbf{x})\right] 
   \right\|
\\ 
   =& 
   \left\|
   \left[g_{\hat{y}}(\mathbf{x} + \delta \mathbf{x}) - g_{\hat{y}}(\mathbf{x})\right] + 
   \left[g_{k}(\mathbf{x}) - g_{k}(\mathbf{x} + \delta \mathbf{x})\right] 
   \right\| 
\\ 
\leq& \left\|(g_{\hat{y}}(\mathbf{x} + \delta \mathbf{x})-g_{\hat{y}}(\mathbf{x}))\right\|+ \left\| (g_{k}(\mathbf{x} + \delta \mathbf{x})-g_{k}(\mathbf{x}))\right\|
\\
\leq& 2\left\|\delta \mathbf{y}\right\|
\end{aligned}
\end{equation}
Therefore, we obtain the upper bound of the local Lipschitz constant,
\begin{equation}
\label{eq:upper_bound} 
    L_{q,\mathbf{x}}^{k} = 
    \underset{\|\delta \mathbf{x}\|\leq {r}}{\rm sup} 
    \frac{\|h(\mathbf{x} + \delta \mathbf{x}) - h(\mathbf{x})\|}{\|\delta \mathbf{x}\|} 
    \leq 
    2\frac{\left\|\delta\mathbf{y}\right\|} {\left\|\delta \mathbf{x}\right\|}
\end{equation}
By combining Eq.\eqref{eq:pattern} and Eq.\eqref{eq:upper_bound}, it leads to~\eqref{eq:Lipschitz_condition}.
\end{proof}
\end{theorem}

It can be seen from Eq.\eqref{eq:Lipschitz_condition}, building an extremely sparse neural network means a smaller value of $\left\| \mathbf{W}\right\|$. At the same time, the value of the condition number will sharply increase. This satisfies our expectation that a highly sparse network reduces the rank of the weight matrix, and tends to fall into ill-conditioned weight space, despite the sparsity of the network equally minimizing the local Lipschitz constant. They are linearly correlated and the $L_{q,\mathbf{x}}^{k}$ cannot keep shrinking due to the capacity and the depth of network~\cite{bubeck2021universal,bubeck2021law}. 
According to Eq.\eqref{condition_number} and Eq.\eqref{eq:Lipschitz_condition}, we can draw a conclusion that adequate sparsity of the network can improve the robustness due to limiting the local Lipschitz constant and thus obtaining a larger perturbation boundary. Moreover, a moderate condition number of the weight matrix is also indispensable. 

\subsection{Scale-Invariant Condition Number Constraint}
A central question in training a sparse network is how to reserve the neuron connections which can promote robustness. 
Integrating the adversarial training technique into the pruning procedure is not enough.
As it has been showed from Thm.\ref{Theorem:3.2} that exist $\forall q \in \left\{1,2\right\}$ for $\left\|\mathbf{W}_{j}\right\|_{p}$ can limit the local Lipschitz constant $L_{q,\mathbf{x}}^{k}$. We reckon that promoting the shrinkage of non-zero elements in the weight matrix towards zero elements is equally important to minimize the $L_{q,\mathbf{x}}^{k}$ value. 
Thus, we impose constraints through Frobenius regularization to limit the Local
Lipschitz as follows:
\begin{equation}
\label{equivalent_l2norm}
\begin{aligned}
\mathcal{L}_{\textit{CC}} = 
\sum_{l=1}^{L}
\left(\log \big(\tau + \|\mathbf{W}_l\|_F^2 \big)\right)
\end{aligned}
\end{equation}
Where $L$ is the size of linear layers, and $0<\tau \ll 1$ being the small smoothing factor, which avoids exploding or vanishing gradients during back-propagation, we set $\tau = 1e-4$ in our experiments. The logarithmic function $\log{(\cdot)}$ satisfies:
\begin{equation}
\frac{\partial\widetilde{\mathbf{W}}_{l}}{\partial\widetilde{w}_{l,ij}}
= \frac{\partial(\mu_{l}\mathbf{W}_{l})}{\partial(\mu_{l} w_{l,ij})}
= \frac{\partial\mathbf{W}_{l}}{\partial w_{l,ij}}.
\end{equation}
Here, $\widetilde{\mathbf{W}}_{l} = \mu_{l}\mathbf{W}_{l}$, $\widetilde{\mathbf{b}}_{l} = \mu_{l}\mathbf{b}_{l}$. For the positive homogeneity function ($\sigma(\widetilde{\mathbf{W}}_{l}\mathbf{x}_{l-1} + {\widetilde{\mathbf{b}}}_{l}) = \mu_{l} \cdot \sigma(\mathbf{W}_{l}\mathbf{x}_{l-1} + \mathbf{b}_{l})$ when $a > 0$), 
since penalizing the intrinsic norms of the weight matrix is generally ineffective \cite{liu2021improve}, so we choose $\log{(\cdot)}$ to measure the Scale-invariant $L_{2}$ \cite{wei2022learning}.

Meanwhile, we observe that it is not enough to train a robust model relying on sparse constraint alone, due to the reason that extremely sparse network causes the matrix to be non-full rank, i.e., leading to higher condition number $\kappa\left(\mathbf{W}\right)$ in the sparse step. 
Orthogonal neural network~\cite{bansal2018can} forces the condition number of weight matrices to be close to $1$ through spectral constraint.
Let $\{\sigma_i\}_{i = 1}^k, k = \min\{a, b\}$ denote the singular values of $a$ weight matrix, $\mathbf{W}_{l} \in \mathbb{R}^{a\times b}$ arranged in descending order of magnitude. and $\sigma_{\max}(\mathbf{W}_l)^2$ denotes the largest one. It is known that $\|\mathbf{W}_l\|_F^2 = {\sum_{i = 1}^{k}\sigma_i^2} \geq\sigma_{\max}(\mathbf{W}_l)^2 $, thus, the regularization term in Eq.~\eqref{equivalent_l2norm} could also prevent condition number from being too large.

\begin{figure}
    \centering
    \includegraphics[width=\linewidth]{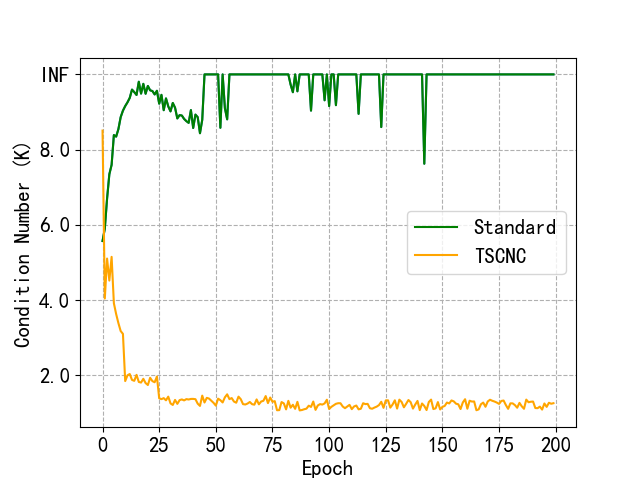}
    \caption{The variation of the condition number in the process of adversarial training with VGG-16, where 'Standard' indicates the pretaining network.}
    \label{fig:cnumber}
\end{figure}

\begin{algorithm}
\SetAlgoLined
\caption{TSCNC}
\label{alg:tscnc}
\KwIn{reference model weights $\mathbf{W}_l (0 \leq l \leq L)$; sparsity $p\%$; input samples $(x,y) \sim \mathcal{D}$}
\KwOut{sparse model weights $\mathbf{W}_l$}

\For{each sample $(x, y) \sim \mathcal{D}$}{
    $x_{\text{adv}} \leftarrow \text{attack}(x, y)$\;
    \For{each layer $l$}{
     $\Delta \mathbf{W} \leftarrow -\mathbf{W} \odot \Delta \mathbf{Z}$\\
        \uIf{layer is convolution}{
            $\Delta \mathcal{L}_{\textit{CE}} \leftarrow \left\{\sum_i \left[ \nabla_{z_i} \mathcal{L} \times \mathbf{a}_i^\top \right] \right\} \Delta \mathbf{W}$\;
        }
        \uElseIf{layer is fully connected}{
            $\Delta \mathcal{L}_{\textit{CE}} \leftarrow \{\nabla_z \mathcal{L}\} \mathbf{a}^\top \Delta \mathbf{W}$\;
        }
    }
}
Sort $\Delta \mathcal{L}_{\textit{CE}}$ and set mask matrix $\mathbf{Z}$\;

\For{each epoch}{
    $x_{\text{adv}} \leftarrow \text{attack}(x, y)$\;
    \For{each layer $l$}{
        $\widetilde{\mathbf{W}}_l \leftarrow \mathbf{W}_l \odot \mathbf{Z}_l$\;
        $\mathbf{W}_l \leftarrow \arg\min\limits_{\widetilde{\mathbf{W}}_l} \{\mathcal{L}_{\textit{E}} + \lambda \mathcal{L}_{\textit{CC}}\}$\;
    }
}
\Return $\mathbf{W}_l$\;
\end{algorithm}

\begin{table*}[htbp]
\centering
\small
\resizebox{\textwidth}{!}{%
\begin{tabular}{c|l|ccc|ccc|ccc|ccc}
\toprule
 & \multicolumn{1}{c|}{} & \multicolumn{6}{c}{\textbf{CIFAR-10}} & \multicolumn{6}{c}{\textbf{CIFAR-100}} \\ \cmidrule(lr){1-2} \cmidrule(lr){3-8} \cmidrule(lr){9-14}
\multirow{2}{*}{Model} & \multicolumn{1}{c|}{Sparsity} & \multicolumn{3}{c}{90\%} & \multicolumn{3}{c}{95\%} & \multicolumn{3}{c}{90\%} & \multicolumn{3}{c}{95\%} \\ \cmidrule(lr){2-2} \cmidrule(lr){3-5} \cmidrule(lr){6-8} \cmidrule(lr){9-11} \cmidrule(lr){12-14} 
 & \multicolumn{1}{c|}{Method} & HYDRA & MAD & TSCNC & HYDRA & MAD & TSCNC & HYDRA & MAD & TSCNC & HYDRA & MAD & TSCNC \\ \midrule
\multirow{5}{*}{\shortstack{VGG-16 \\ \textit{(14.7M)}}} & 
Clean & 76.40 & 81.40 & \textbf{81.94} & 74.50 & 71.60 & \textbf{81.30} & 52.56 & 48.71 & \textbf{54.20} & 50.78 & 44.77 & \textbf{53.57} \\
 & FGSM & 58.94 & 57.07 & \textbf{60.17} & 55.91 & 54.81 & \textbf{58.74} & 32.40 & 35.71 & \textbf{38.97} & 30.77 & 33.05 & \textbf{36.49} \\
 & PGD & 49.58 & 51.08 & \textbf{53.27} & 48.73 & 48.44 & \textbf{51.87} & 25.32 & 26.61 & \textbf{29.10} & 21.15 & 22.64 & \textbf{25.97} \\
 & SA & 53.24 & 53.61 & \textbf{55.80} & 50.58 & 50.46 & \textbf{53.36} & 34.70 & 33.34 & \textbf{36.02} & 32.91 & 31.47 & \textbf{34.23} \\
 & AA & 42.55 & 43.12 & \textbf{49.49} & 40.19 & 40.62 & \textbf{46.43} & 19.87 & 20.28 & \textbf{24.95} & 18.50 & 19.02 & \textbf{23.83} \\ \midrule
\multirow{5}{*}{\shortstack{WRN-28-10 \\ \textit{(36.5M)}}} & 
Clean & 87.40 & 88.00 & \textbf{89.64} & --.-- & 72.36 & \textbf{84.90} & 62.50 & 48.27 & \textbf{62.82} & 55.47 & 43.84 & \textbf{57.78} \\
& PGD & 53.27 & 54.20 & \textbf{57.63} & 52.21 & 53.85 & \textbf{56.32} & 28.63 & 29.64 & \textbf{32.97} & 28.60 & 27.41 & \textbf{30.71} \\
 & APGD & 49.47 & 51.36 & \textbf{54.91} & 47.78 & 49.21 & \textbf{53.67} & 26.79 & 28.61 & \textbf{31.73} & 26.28 & 26.57 & \textbf{30.41} \\
 & CW$_{\infty}$ & 51.29 & 53.44 & \textbf{57.01} & 50.55 & 51.45 & \textbf{55.97} & 25.53 & 27.60 & \textbf{29.43} & 23.89 & 24.11 & \textbf{28.41} \\
 & AA & 49.16 & 49.50 & \textbf{54.09} & 47.46 & 47.55 & \textbf{53.54} & 24.63 & 24.48 & \textbf{27.34} & 21.30 & 21.64 & \textbf{25.78} \\ \bottomrule
\end{tabular}%
}
\caption{Comparing the adversarial robustness accuracy on small-scaled datasets CIFAR-10 and CIFAR-100 with VGG-16 and WideResNet-28-10. We control the sparsity of models to 90\% and 95\%, the bold indicates the best performance at the same sparsity under the same attack method. See results of VGG-11, ResNet-18 in the \textit{Supplementary Materials}.}
\label{tab:1}
\end{table*}

\subsection{Transformed Sparse Model}
In training networks under a sparse model with high levels of pruning rates, one practice is to remove weights that have less impact on accuracy. We achieve this by first employing an adaptive masks matrix directly multiplying the weight matrix, then marking the corresponding model parameter whether to remove or not. 

Let a mask matrix $\mathbf{Z}\in \mathbb{R}^{m\times n}$ in which the elements $\mathbf{Z}_{i,j} \in \{0,1\}$ control the elements of weight matrix $\mathbf{W}\in \mathbb{R}^{m\times n}$ whether to remove. Then we can reformulate the weight matrix $\mathbf{W}$ as $\widetilde{\mathbf{W}}= \mathbf{W}\odot\mathbf{Z}$ and the min-max empirical risk function in models $h\left( \cdot \right) $ can be rewritten as:
\begin{equation}
\mathcal{L}_{\textit{E}} = 
\min_{\widetilde{\mathbf{W}}} {\mathbb{E}_{p}}_{(\mathbf{x}_{i},y_{i})\sim \mathcal{D}}
\left[
\max_{\left\|\delta\right\|_{p}\leq\gamma } \mathcal{L}_{\textit{CE}}(h(\mathbf{x}_{i}+\delta ;\widetilde{\mathbf{W}}), y_{i})
\right]
\end{equation}
where, $\delta$ denotes the adversarial perturbation added to the input tensor; $\left\| \cdot \right\|_{p}$ denotes the $\ell_{p}$ norm of adversarial perturbation and we set $p=\infty $ in experiments, $\mathcal{L}_{\textit{CE}}$ represents the Cross Entropy Loss function.
For training sparse networks, we selectively prune model parameters corresponding to small loss change with $\mathcal{L}_{\textit{CE}}$. So we can use Taylor expansion to express the change of loss function as:
\begin{equation}
\Delta \mathcal{L}_{\textit{CE}} =
\frac{\partial \mathcal{L}_{\textit{CE}}}{\partial \mathbf{w}} \Delta \mathbf{w} + \frac{1}{2} \Delta \mathbf{w}^{\top} \mathbf{H} \Delta \mathbf{w}
\label{eq_losschange}
\end{equation}
where $\mathbf{H}$ denotes the \textit{Hessian matrix} computed as the second-order derivatives of the loss function. Generally, the difference $\Delta \mathbf{w}$ gets small enough, and the second-order gradient is close to zero. Thus, we set a variable $\beta$ which is determined by the sparsity rate.
Once the absolute value of $\Delta \mathcal{L}_{\textit{CE}}$ is less than $\beta$, then $z_{i,j}$ is assigned to zero value, which explicitly means that the parameter of $w_{i,j}$ is selected to remove. Conversely, if the absolute value is larger than $\beta$, then $z_{i,j}$ is assigned to one value,  which implies that we will reserve $w_{i,j}$ in the whole training process. 

It is easy to acquire the value of $\Delta\mathbf{w}$ by calculating the difference between before and after the back-propagation. There is another important issue to computing first-order information of loss function, the following Propositions describe the method of calculating the gradient of the weight matrix for the convolution layer and the fully-connected layer.
\begin{proposition} 
\label{prosition3.3}
For $l^{th}$ convolution layer, firstly, we define weight matrix $\mathbf{W} \in \mathbb{R}^{c_{\textit{out}} \times c_{\textit{in}}k^2}$, activation map $\mathbf{A}\in \mathbb{R}^{c_{\textit{in}} \times s_a}$, and layer output $\mathbf{Z} \in \mathbb{R}^{c_{\textit{out}} \times s_z}$ such that it satisfies $\mathbf{Z}=\mathbf{W}\mathbf{A}$, 
$\nabla_\mathbf{W}\mathcal{L} = \sum_i \left[ \nabla_{\mathbf{Z}_i} \mathcal{L} \times \mathbf{A}_i^{\top} \right]$
where $i$ expresses spatial index. 
Note that $k$ is the weight kernel size, $c_{\textit{in}}$ is channel number of $l^{th}$ layer, $c_{\textit{out}}$ is channel number of $(l+1)^{th}$ layer, and $s_a$ is spatial size of activation map $a$, $s_z$ is spatial size of layer output $z$.
\end{proposition}

\begin{proposition}
\label{prosition3.4}
For $l^{th}$ fully-connected layer, firstly, we define weight matrix $\mathbf{W} \in \mathbb{R}^{m \times n}$, (activated) layer input $\mathbf{a}\in \mathbb{R}^m$, and layer output $\mathbf{z} \in \mathbb{R}^n$ such that it satisfies $\mathbf{z} = \mathbf{W}^{\top} \mathbf{a}, \nabla_\mathbf{W} \mathcal{L} = \mathbf{a}\{\nabla_{\mathbf{Z}} \mathcal{L}\}^{\top}$. Note that m and n denote the total number of nodes in each $l^{th}$ and $(l + 1)^{th}$ layer.
\end{proposition}

According Proposition~\ref{prosition3.3} and \ref{prosition3.4}, we expand the equation \cref{eq_losschange} for $l^{th}$ convolution layer and fully-connected layer respectively as follows:
\begin{equation}
\label{eq_convolution}
\begin{aligned}
\mathrm{Conv:}\quad & \Delta \mathcal{L}_{\textit{CE}} 
= \left\{\sum_i \nabla_{\mathbf{z}_i} \mathcal{L}\mathbf{a}_i^{\top} \right\} \Delta \mathbf{W} \\
\mathrm{FC:}\quad   & \Delta \mathcal{L}_{\textit{CE}} 
= \{\nabla_{\mathbf{z}} \mathcal{L}\}\mathbf{a}^{\top} \Delta \mathbf{W}
\end{aligned}
\end{equation}

$\Delta \mathcal{L}_{\textit{CE}}$ is a key idea of our pruning methodology, because it provides us with internal information about the weight matrix $\mathbf{W}$, letting us know which model parameters are useful or useless factors, and we can determine which model parameters can be pruned without weakening the adversarial robustness.
If the model parameters with larger loss change with $\Delta \mathcal{L}_{\textit{CE}}$, We can conclude that even if small changes occur, they can easily affect the model's prediction of adversarial examples. Thus, they can be considered as factors for gaining robust knowledge through adversarial training.
In the opposite case, we argue that the model parameters corresponding to small $\Delta \mathcal{L}_{\textit{CE}}$ are not so prominent in the adversarial example that their effects can simply be ignored. Therefore, If we preset the model pruning rate, we can easily prune the model parameters, just eliminating them in order of low adversarial $\Delta \mathcal{L}_{\textit{CE}}$.

\begin{table}[htbp]
\centering
\resizebox{0.9\columnwidth}{!}{%
\begin{tabular}{l|c|l|ccc}
\toprule
\multicolumn{6}{c}{\textbf{Tiny-Imagenet}} \\ \midrule
Model & Sparsity & Method & HYDRA & MAD & TSCNC \\ \midrule
\multirow{12}{*}{\shortstack{VGG-16 \\ \textit{(14.7M)}}} & \multirow{6}{*}{90\%} 
& Clean & --.-- & \underline{47.00} & \textbf{47.17} \\ 
& & FGSM & 17.83 & 18.90 & \textbf{23.47} \\
& & PGD & 16.21 & 16.35 & \textbf{19.97} \\
& & CW$_{\infty}$ & 13.44 & 13.58 & \textbf{16.78} \\
& & APGD & 13.95 & 14.56 & \textbf{17.80} \\
& & AA & 11.80 & 12.44 & \textbf{15.57} \\ \cmidrule{2-6}
& \multirow{6}{*}{95\%} 
& Clean & --.-- & 24.72 & \textbf{45.21} \\ 
& 
& FGSM & 16.34 & 16.90 & \textbf{21.07} \\
& & PGD & 14.68 & 14.35 & \textbf{18.86} \\
& & CW$_{\infty}$ & 12.33 & 13.10 & \textbf{18.84} \\
& & APGD & 12.02 & 12.89 & \textbf{16.25} \\
& & AA & 11.00 & 11.41 & \textbf{14.39} \\ \bottomrule
\end{tabular}
}
\caption{Comparing the adversarial robustness accuracy on Tiny-ImageNet dataset with VGG-16. Underline: from MAD paper.}
\label{tab:2}
\end{table}

\subsection{Training Robust Model with Sparse Constraint}
We focus on training a robust neural network at high sparsity. We build theoretical relationships among the sparsity, the local Lipschitz constant, and the condition number of neural networks. Thus, the constraints for condition number within the framework of sparse DNNs with limiting the $L_{q,\mathbf{x}}^{k}$ can exactly improve the robustness against various perturbations. By combining the constraints and sparse constraint discussed above, we construct the following cost function to jointly learn extremely sparse and robust models:
\begin{align}
    \mathcal{R}(\widetilde{\mathbf{W} })=\mathcal{L}_{\textit{E}}(\widetilde{\mathbf{W} })+ \lambda \mathcal{L}_{\textit{CC}}
\end{align}
where the hyperparameter  $\lambda > 0$ controls the influence of the regularizers on the final solution.
In this work, we describe~\cref{alg:tscnc} to explain TSCNC in detail.

\section{Experimental Results}

\subsection{Implementation Settings}

We here introduce adversarial attack methods and hyperparameter settings.

\subsubsection{Adversarial Attack}

We verify the effectiveness of our model under different adversarial attack methods, such as Fast Gradient Sign Method (FGSM) attack~\cite{goodfellow2014explaining}, the Projected Gradient Descent (PGD) attack~\cite{madry2017towards}, Auto PGD (APGD)~\cite{croce2020reliable}, Square Attack (SA)~\cite{andriushchenko2020square}, CW$_{\infty}$~\cite{carlini2017towards} and Auto Attack (AA)~\cite{croce2020reliable}. 
We set the PGD attack and FGSM with the perturbation magnitude $\epsilon =8/255$, steps $t=10$, and step size $\alpha=2/255$ for $\ell_{\infty }$ norm attack. The SA is a black-box attack without any internal knowledge of the targeted model, and the number of queries is set to 100. We implement APGD with cross-entropy loss, and it has 30 steps with random starts and momentum coefficient $\rho =0.75$. We use CW$_{\infty}$ attack with the same perturbation budget as the PGD attack.

\subsubsection{Performance Metrics and Hyperparameters} 
For all approaches, we adversarially train models by using TRADES-AT~\cite{zhang2019theoretically} with $\epsilon=8/255$, perturb steps=$2/255$ and distance=$l_{\infty }$ for adversarial loss. The proposed method is compared with recent advanced baselines: {\bf HYDRA}~\cite{sehwag2020hydra}, {\bf MAD}~\cite{lee2022masking}, and for baselines, the optimal hyperparameters are consistent with those given in the paper. 
We set the batch size to $128$ and the epoch to $200$ for all methods. In our method, we train the sparse model from scratch using SGD with momentum $0.9$, weight decay $0.0005$, initial learning rate $0.1$ and take a step scheduler to lower the learning rate by $0.1$ times on each $30$ and $45$ epochs.

\subsection{Results and Discussion} 

We report the results of our comparative evaluation with HYDRA~\cite{sehwag2020hydra} and MAD~\cite{lee2022masking} 
Besides performance on adversarial samples, our performance on clean samples also outperforms existing methods. 

Our method outperforms existing works in all experiments.
As shown in~\cref{90sparsity}, the distribution of parameter values trained by TSCNC is closer to zero than HYDRA. Our method tends to retain the parameters with small values but high adversarial robustness.
TSCNC is $4.3\%$ and $4.06\%$ higher than HYDRA for CIFAR-10 and CIFAR-100 each on average of two networks.
This is because the hidden layers and fully connected layers of the model pruned by HYDRA mostly fall into non-full rank weight matrix space as shown in~\cref{comparsion_acc}, affecting robustness when encountering stronger adversarial attacks. 

TSCNC has $1.55\%$ and $1.83\%$ degradation for CIFAR-10 and CIFAR-100 respectively on an average of two networks, whereas MAD has $2.34\%$ and $2.54\%$ degradation respectively. MAD has no change in the value of the parameter, simply determining whether to pruned a parameter based on the loss change. It can be seen in~\cref{90sparsity,95sparsity}, that TSCNC constraints the weight distribution and tends to retain smaller values of parameters than MAD.  

\subsection{Ablation Study}

We analyse the effectiveness of the proposed constraint in \cref{sec:431}, and the impact on robustness under different sparsity in \cref{sec:432}.

\subsubsection{Regarding Constraint}
\label{sec:431}

\paragraph{Constraint}
Since we argue that $\mathcal{L}_{\textit{CC}}$ are essential to adverse resilience, we evaluate the effect of this component on CIFAR-10 using WRN-28-10, as shown in~\cref{ablation}. 
We conduct two experiments between just sparsity training and using conditional number constraint $\mathcal{L}_{\textit{CC}}$. 

As observed in~\cref{ablation}, $\mathcal{L}_{\textit{CC}}$ is particularly beneficial for adversarial generalization of the model under high sparsity, being able to prevent a matrix from falling into the ill-conditioned weight space. Thus, the inclusion of $\mathcal{L}_{\textit{CC}}$ terms allows TSCNC to perform well. Moreover, as shown in~\cref{fig:cnumber}, TSCNC always ensures that the condition number of the model is in a lower value range during training. 

\begin{table}[h]
\centering
\resizebox{\linewidth}{!}{
\begin{tabular}{c|cccc}
\toprule
\multirow{2}{*}{Sparsity} & \multicolumn{4}{c}{Attack Type (w.o./w. $\mathcal{L}_{\textit{CC}}$)} \\\cmidrule(lr){2-5}
& PGD & CW$_{\infty}$ & APGD & AA \\\midrule
90\% & 55.25/\textbf{57.63} & 54.69/\textbf{57.01} & 50.89/\textbf{54.91} & 49.92/\textbf{54.09} \\
95\% & 51.37/\textbf{56.32} & 49.76/\textbf{55.97} & 46.88/\textbf{53.67} & 45.61/\textbf{53.54} \\
\bottomrule
\end{tabular}
}
\caption{The results of ablation study on the presence of $\mathcal{L}_{\textit{CC}}$ in TSCNC (WRN-28-10) on CIFAR-10 dataset.}
\label{ablation}
\end{table}

\begin{figure}
    \centering
    \includegraphics[width=\linewidth]{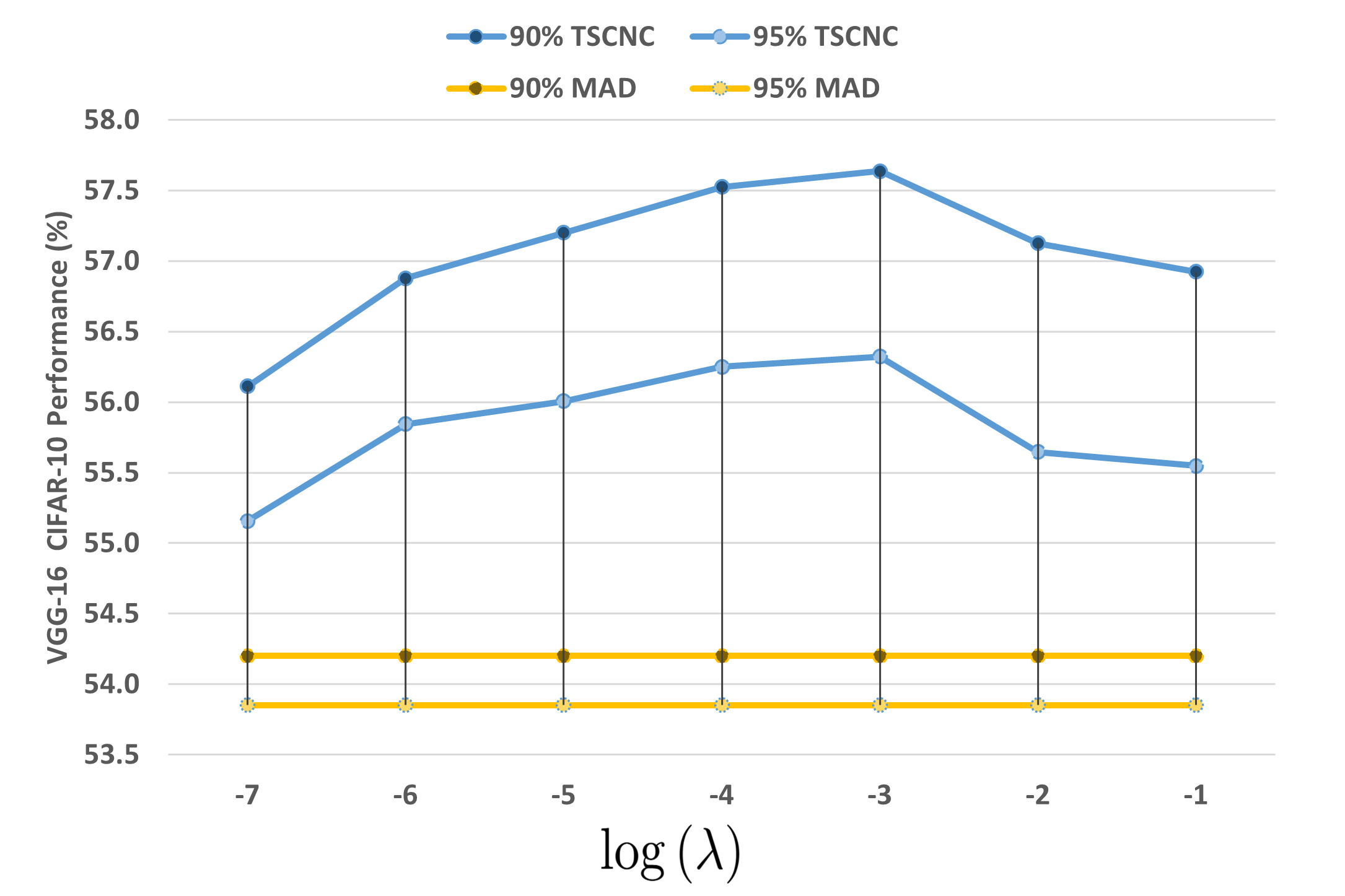}
    \caption{The impact of different value of $\lambda$ on robust accuracy, the pruning ratio is set to $90\%$ and $95\%$, respectively.}
    \label{fig:hyperparameter}
\end{figure}

\paragraph{Hyperparameter}

$\mathcal{L}_{\textit{CC}}$ has a single hyperparameter $\lambda$, which determines the magnitude of the constraint of the condition number, and in turn, controls the distribution of the parameters in the weight matrix. It is necessary to investigate how $\lambda$ controls the robustness of DNNs. \cref{fig:hyperparameter} shows the impact of $\lambda$ under the PGD attack on the CIFAR-10 dataset, showing that with the increase of $\lambda$, the precision of the model is on the rise, this is mainly because regularization can limit the local Lipschitz constant when the input produces tiny perturbation. However, when $\lambda$ continues to increase, the overall loss value will be too large, which makes the gradient disappearance easy to occur in the training process and makes it difficult to converge. we finally determined $ \lambda = 0.001$ with better performance comparatively.

\subsubsection{Regarding Sparsity and Robustness}
\label{sec:432}

\begin{figure}
    \centering
    \includegraphics[width=\linewidth]{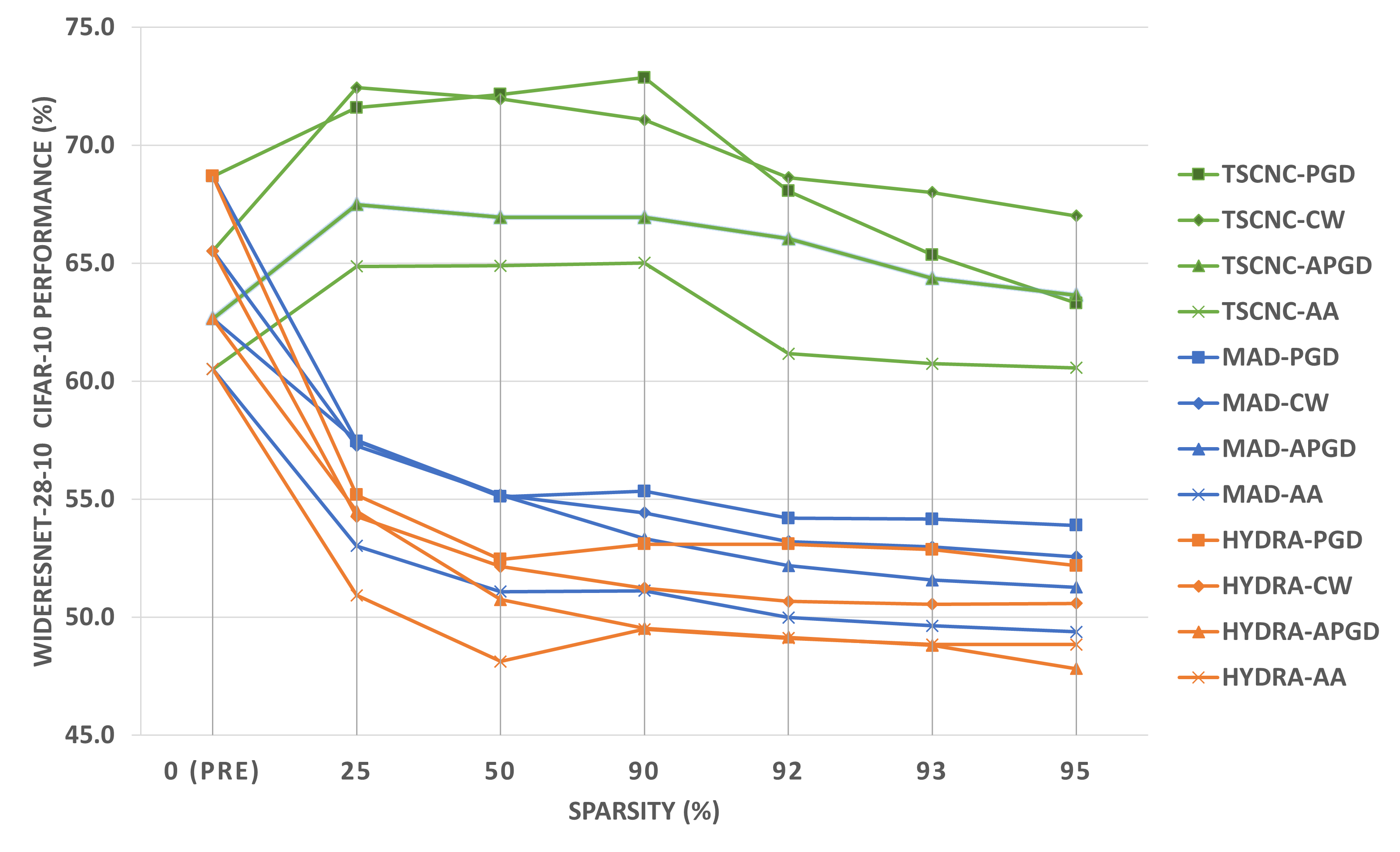}
    \caption{The results at different sparsity under various attack methods, we select WRN-28-10 with PGD adversarial training on CIFAR-10 compared with HYDRA and MAD.}
    \label{fig:3_attack_fig}
\end{figure}

We here look into the robustness with variation of sparsity. 
We compare TSCNC with other methods under 4 types of attack, using WideResNet-28-10 on CIFAR-10. 
As shown in Fig.~\ref{fig:3_attack_fig}, we plot the robust accuracy curves. On the one hand, the results confirm that the robustness of the model can benefit from moderate sparsity, e.g., the robust accuracy of TSCNC rises by 4.08\% (68.68\% to 72.76\%) from 0\% to 90\% sparsity under PGD attack. The baseline methods drop at varying degrees.

\section{Conclusion}

In this work, we theoretically demonstrate the relationship between the sparsity, the local Lipschitz constant, and the condition number of deep neural networks. Based on such findings, we propose novel joint constraints to balance the sparsity and robustness of DNNs. The proposed method combines smoothing distribution and differentiable constraints, to reduce the condition number of the weight matrix and avoid falling into ill-conditioned space. Various experiments are conducted on different datasets and attack methods to verify the effectiveness of the proposed method.

{
    \small
    \bibliographystyle{ieeenat_fullname}
    \bibliography{main}

\begin{thebibliography}{39}
\providecommand{\natexlab}[1]{#1}
\providecommand{\url}[1]{\texttt{#1}}
\expandafter\ifx\csname urlstyle\endcsname\relax
  \providecommand{\doi}[1]{doi: #1}\else
  \providecommand{\doi}{doi: \begingroup \urlstyle{rm}\Url}\fi

\bibitem[Andriushchenko et~al.(2020)Andriushchenko, Croce, Flammarion, and Hein]{andriushchenko2020square}
Maksym Andriushchenko, Francesco Croce, Nicolas Flammarion, and Matthias Hein.
\newblock Square attack: a query-efficient black-box adversarial attack via random search.
\newblock In \emph{European Conference on Computer Vision (ECCV)}, pages 484--501. Springer, 2020.

\bibitem[Anwar et~al.(2017)Anwar, Hwang, and Sung]{anwar2017structured}
Sajid Anwar, Kyuyeon Hwang, and Wonyong Sung.
\newblock Structured pruning of deep convolutional neural networks.
\newblock \emph{ACM Journal on Emerging Technologies in Computing Systems (JETC)}, 13\penalty0 (3):\penalty0 1--18, 2017.

\bibitem[Bansal et~al.(2018)Bansal, Chen, and Wang]{bansal2018can}
Nitin Bansal, Xiaohan Chen, and Zhangyang Wang.
\newblock Can we gain more from orthogonality regularizations in training deep networks?
\newblock \emph{Advances in Neural Information Processing Systems (NeurIPS)}, 31, 2018.

\bibitem[Biggio et~al.(2013)Biggio, Corona, Maiorca, Nelson, {\v{S}}rndi{\'c}, Laskov, Giacinto, and Roli]{biggio2013evasion}
Battista Biggio, Igino Corona, Davide Maiorca, Blaine Nelson, Nedim {\v{S}}rndi{\'c}, Pavel Laskov, Giorgio Giacinto, and Fabio Roli.
\newblock Evasion attacks against machine learning at test time.
\newblock In \emph{European Conference on Machine Learning and Knowledge Discovery in Databases}, pages 387--402. Springer, 2013.

\bibitem[Bubeck and Sellke(2021)]{bubeck2021universal}
S{\'e}bastien Bubeck and Mark Sellke.
\newblock A universal law of robustness via isoperimetry.
\newblock \emph{Advances in Neural Information Processing Systems (NeurIPS)}, 34:\penalty0 28811--28822, 2021.

\bibitem[Bubeck et~al.(2021)Bubeck, Li, and Nagaraj]{bubeck2021law}
S{\'e}bastien Bubeck, Yuanzhi Li, and Dheeraj~M Nagaraj.
\newblock A law of robustness for two-layers neural networks.
\newblock In \emph{Conference on Learning Theory}, pages 804--820. PMLR, 2021.

\bibitem[Carlini and Wagner(2017)]{carlini2017towards}
Nicholas Carlini and David Wagner.
\newblock Towards evaluating the robustness of neural networks.
\newblock In \emph{2017 IEEE Symposium on Security and Privacy (SP)}, pages 39--57, 2017.

\bibitem[Chen et~al.(2022)Chen, Zhang, Zhang, Chang, Liu, Chen, and Wang]{chen2022linearity}
Tianlong Chen, Huan Zhang, Zhenyu Zhang, Shiyu Chang, Sijia Liu, Pin-Yu Chen, and Zhangyang Wang.
\newblock Linearity grafting: Relaxed neuron pruning helps certifiable robustness.
\newblock In \emph{International Conference on Machine Learning (ICML)}, pages 3760--3772. PMLR, 2022.

\bibitem[Cheng et~al.(2024)Cheng, Zhang, and Shi]{PAMI24Pruning2}
Hongrong Cheng, Miao Zhang, and Javen~Qinfeng Shi.
\newblock A survey on deep neural network pruning: Taxonomy, comparison, analysis, and recommendations.
\newblock \emph{IEEE Transactions on Pattern Analysis and Machine Intelligence}, 46\penalty0 (12):\penalty0 10558--10578, 2024.

\bibitem[Cosentino et~al.(2019)Cosentino, Zaiter, Pei, and Zhu]{cosentino2019search}
Justin Cosentino, Federico Zaiter, Dan Pei, and Jun Zhu.
\newblock The search for sparse, robust neural networks.
\newblock \emph{arXiv preprint arXiv:1912.02386}, 2019.

\bibitem[Croce and Hein(2020)]{croce2020reliable}
Francesco Croce and Matthias Hein.
\newblock Reliable evaluation of adversarial robustness with an ensemble of diverse parameter-free attacks.
\newblock In \emph{International Conference on Machine Learning (ICML)}, pages 2206--2216. PMLR, 2020.

\bibitem[Denil et~al.(2013)Denil, Shakibi, Dinh, Ranzato, and De~Freitas]{denil2013predicting}
Misha Denil, Babak Shakibi, Laurent Dinh, Marc'Aurelio Ranzato, and Nando De~Freitas.
\newblock Predicting parameters in deep learning.
\newblock \emph{Advances in Neural Information Processing Systems (NeurIPS)}, 26, 2013.

\bibitem[Goodfellow et~al.(2014)Goodfellow, Shlens, and Szegedy]{goodfellow2014explaining}
Ian~J Goodfellow, Jonathon Shlens, and Christian Szegedy.
\newblock Explaining and harnessing adversarial examples.
\newblock In \emph{International Conference on Learning Representations (ICLR)}, 2014.

\bibitem[Gui et~al.(2019)Gui, Wang, Yang, Yu, Wang, and Liu]{gui2019model}
Shupeng Gui, Haotao Wang, Haichuan Yang, Chen Yu, Zhangyang Wang, and Ji Liu.
\newblock Model compression with adversarial robustness: A unified optimization framework.
\newblock \emph{Advances in Neural Information Processing Systems (NeurIPS)}, 32, 2019.

\bibitem[Guo et~al.(2016)Guo, Yao, and Chen]{guo2016dynamic}
Yiwen Guo, Anbang Yao, and Yurong Chen.
\newblock Dynamic network surgery for efficient dnns.
\newblock \emph{Advances in Neural Information Processing Systems (NeurIPS)}, 29, 2016.

\bibitem[Guo et~al.(2018)Guo, Zhang, Zhang, and Chen]{guo2018sparse}
Yiwen Guo, Chao Zhang, Changshui Zhang, and Yurong Chen.
\newblock Sparse dnns with improved adversarial robustness.
\newblock \emph{Advances in Neural Information Processing Systems (NeurIPS)}, 31, 2018.

\bibitem[Han et~al.(2015)Han, Pool, Tran, and Dally]{han2015learning}
Song Han, Jeff Pool, John Tran, and William Dally.
\newblock Learning both weights and connections for efficient neural network.
\newblock \emph{Advances in Neural Information Processing Systems (NeurIPS)}, 28, 2015.

\bibitem[Han et~al.(2016)Han, Mao, and Dally]{han2015deep}
Song Han, Huizi Mao, and William~J Dally.
\newblock Deep compression: Compressing deep neural networks with pruning, trained quantization and huffman coding.
\newblock In \emph{International Conference on Learning Representations (ICLR)}, 2016.

\bibitem[He and Xiao(2024)]{PAMI24Pruning}
Yang He and Lingao Xiao.
\newblock Structured pruning for deep convolutional neural networks: A survey.
\newblock \emph{IEEE Transactions on Pattern Analysis and Machine Intelligence}, 46\penalty0 (5):\penalty0 2900--2919, 2024.

\bibitem[Hein and Andriushchenko(2017)]{hein2017formal}
Matthias Hein and Maksym Andriushchenko.
\newblock Formal guarantees on the robustness of a classifier against adversarial manipulation.
\newblock \emph{Advances in Neural Information Processing Systems (NeurIPS)}, 30, 2017.

\bibitem[Krizhevsky et~al.(2009)Krizhevsky, Hinton, et~al.]{krizhevsky2009learning}
Alex Krizhevsky, Geoffrey Hinton, et~al.
\newblock Learning multiple layers of features from tiny images.
\newblock 2009.

\bibitem[Kukreja et~al.(2006)Kukreja, L{\"o}fberg, and Brenner]{kukreja2006least}
Sunil~L Kukreja, Johan L{\"o}fberg, and Martin~J Brenner.
\newblock A least absolute shrinkage and selection operator (lasso) for nonlinear system identification.
\newblock \emph{IFAC Proceedings Volumes}, 39\penalty0 (1):\penalty0 814--819, 2006.

\bibitem[Le and Yang(2015)]{le2015tiny}
Yann Le and Xuan Yang.
\newblock Tiny imagenet visual recognition challenge.
\newblock \emph{CS 231N}, 7\penalty0 (7):\penalty0 3, 2015.

\bibitem[Lee et~al.(2022)Lee, Kim, and Ro]{lee2022masking}
Byung-Kwan Lee, Junho Kim, and Yong~Man Ro.
\newblock Masking adversarial damage: Finding adversarial saliency for robust and sparse network.
\newblock In \emph{IEEE/CVF Conference on Computer Vision and Pattern Recognition (CVPR)}, pages 15126--15136, 2022.

\bibitem[Liu et~al.(2021)Liu, Cui, and Chan]{liu2021improve}
Ziquan Liu, Yufei Cui, and Antoni~B Chan.
\newblock Improve generalization and robustness of neural networks via weight scale shifting invariant regularizations.
\newblock In \emph{ICML 2021 Workshop on Adversarial Machine Learning}, 2021.

\bibitem[Louizos et~al.(2018)Louizos, Welling, and Kingma]{louizos2017learning}
Christos Louizos, Max Welling, and Diederik~P Kingma.
\newblock Learning sparse neural networks through l\_0 regularization.
\newblock In \emph{International Conference on Learning Representations (ICLR)}, 2018.

\bibitem[Madaan et~al.(2020)Madaan, Shin, and Hwang]{madaan2020adversarial}
Divyam Madaan, Jinwoo Shin, and Sung~Ju Hwang.
\newblock Adversarial neural pruning with latent vulnerability suppression.
\newblock In \emph{International Conference on Machine Learning (ICML)}, pages 6575--6585. PMLR, 2020.

\bibitem[Madry et~al.(2018)Madry, Makelov, Schmidt, Tsipras, and Vladu]{madry2017towards}
Aleksander Madry, Aleksandar Makelov, Ludwig Schmidt, Dimitris Tsipras, and Adrian Vladu.
\newblock Towards deep learning models resistant to adversarial attacks.
\newblock In \emph{International Conference on Learning Representations (ICLR)}, 2018.

\bibitem[Natarajan(1995)]{natarajan1995sparse}
Balas~Kausik Natarajan.
\newblock Sparse approximate solutions to linear systems.
\newblock \emph{SIAM journal on computing}, 24\penalty0 (2):\penalty0 227--234, 1995.

\bibitem[Sehwag et~al.(2020)Sehwag, Wang, Mittal, and Jana]{sehwag2020hydra}
Vikash Sehwag, Shiqi Wang, Prateek Mittal, and Suman Jana.
\newblock Hydra: Pruning adversarially robust neural networks.
\newblock \emph{Advances in Neural Information Processing Systems (NeurIPS)}, 33:\penalty0 19655--19666, 2020.

\bibitem[Sinha et~al.(2018)Sinha, Singh, and Krishnamurthy]{sinha2018neural_EKDD}
Abhishek Sinha, Mayank Singh, and Balaji Krishnamurthy.
\newblock Neural networks in an adversarial setting and ill-conditioned weight space.
\newblock In \emph{Joint European Conference on Machine Learning and Knowledge Discovery in Databases}, pages 177--190. Springer, 2018.

\bibitem[Tsipras et~al.(2018)Tsipras, Santurkar, Engstrom, Turner, and Madry]{tsipras2018robustness}
Dimitris Tsipras, Shibani Santurkar, Logan Engstrom, Alexander Turner, and Aleksander Madry.
\newblock Robustness may be at odds with accuracy.
\newblock \emph{arXiv preprint arXiv:1805.12152}, 2018.

\bibitem[Ullrich et~al.(2017)Ullrich, Meeds, and Welling]{ullrich2017soft}
Karen Ullrich, Edward Meeds, and Max Welling.
\newblock Soft weight-sharing for neural network compression.
\newblock In \emph{International Conference on Learning Representations (ICLR)}, 2017.

\bibitem[Wei et~al.(2022)Wei, Xu, Huang, Lv, Lan, Chen, and Tang]{wei2022learning}
Xian Wei, Yangyu Xu, Yanhui Huang, Hairong Lv, Hai Lan, Mingsong Chen, and Xuan Tang.
\newblock Learning extremely lightweight and robust model with differentiable constraints on sparsity and condition number.
\newblock In \emph{European Conference on Computer Vision (ECCV)}, pages 690--707. Springer, 2022.

\bibitem[Weng et~al.(2018)Weng, Zhang, Chen, Yi, Su, Gao, Hsieh, and Daniel]{weng2018evaluatingl}
Tsui-Wei Weng, Huan Zhang, Pin-Yu Chen, Jinfeng Yi, Dong Su, Yupeng Gao, Cho-Jui Hsieh, and Luca Daniel.
\newblock Evaluating the robustness of neural networks: An extreme value theory approach.
\newblock \emph{International Conference on Learning Representations (ICLR)}, 2018.

\bibitem[Xiao et~al.(2018)Xiao, Tjeng, Shafiullah, and Madry]{xiao2018training}
Kai~Y Xiao, Vincent Tjeng, Nur~Muhammad Shafiullah, and Aleksander Madry.
\newblock Training for faster adversarial robustness verification via inducing relu stability.
\newblock \emph{arXiv preprint arXiv:1809.03008}, 2018.

\bibitem[Ye et~al.(2019)Ye, Xu, Liu, Cheng, Lambrechts, Zhang, Zhou, Ma, Wang, and Lin]{ye2019adversarial}
Shaokai Ye, Kaidi Xu, Sijia Liu, Hao Cheng, Jan-Henrik Lambrechts, Huan Zhang, Aojun Zhou, Kaisheng Ma, Yanzhi Wang, and Xue Lin.
\newblock Adversarial robustness vs. model compression, or both?
\newblock In \emph{IEEE/CVF International Conference on Computer Vision (ICCV)}, pages 111--120, 2019.

\bibitem[Yuval(2011)]{yuval2011reading}
Netzer Yuval.
\newblock Reading digits in natural images with unsupervised feature learning.
\newblock In \emph{Proceedings of the NIPS Workshop on Deep Learning and Unsupervised Feature Learning}, 2011.

\bibitem[Zhang et~al.(2019)Zhang, Yu, Jiao, Xing, El~Ghaoui, and Jordan]{zhang2019theoretically}
Hongyang Zhang, Yaodong Yu, Jiantao Jiao, Eric Xing, Laurent El~Ghaoui, and Michael Jordan.
\newblock Theoretically principled trade-off between robustness and accuracy.
\newblock In \emph{International Conference on Machine Learning (ICML)}, pages 7472--7482. PMLR, 2019.

\end{thebibliography}
}

\clearpage

\appendix

\section{Experimental Environments}

\begin{itemize}
    \item GPU: NVIDIA GeForce RTX 4090
    \item CPU: AMD EPYC 7413
    \item Python Version: 3.11
    \item PyTorch Version: 2.6.0
\end{itemize}

\section{More Details of Condition Number}
Let us consider a general linear transformation $\mathbf{y}=\mathbf{W}\mathbf{x}$, with input-output pairs $(\mathbf{x},\mathbf{y})$ and the weight matrix $\mathbf{W}$. The induced norm of $\mathbf{W}$ measures how much the mapping caused by $\mathbf{W}$ can stretch vectors,
\begin{equation}
    \left\|\mathbf{W}\right\|=\mathop{\max}_{\mathbf{x} \neq 0}\frac{\left\|\mathbf{W}\mathbf{x}\right\|}{\left\| \mathbf{x}\right\|}
\end{equation}

It is also important to consider how much matrix $W$ can shrink vectors. The reciprocal of the minimum stretching can be expressed as the induced norm of the inverse matrix:

\begin{equation}
\begin{aligned}
\left\|\mathbf{W}^{-1}\right\| = \max_{\mathbf{y \neq 
 0}}\frac{\left\|\mathbf{W}^{-1} \mathbf{y}\right\|}{\left\|\mathbf{y}\right\|} 
= \frac{1}{\min_{\mathbf{x \neq 0}}\frac{\left\|\mathbf{W}\mathbf{x}\right\|}{\left\|\mathbf{x}\right\|}}\nonumber
\end{aligned}
\end{equation}
Therefore, an equivalent definition of the Condition Number is as follows:
\begin{equation}
    \kappa(\mathbf{W})= \frac{\sigma_{\max}(\mathbf{W})} {\sigma_{\min}(\mathbf{W})} = \left\|\mathbf{W}\right\|\left\| \mathbf{W}^{-1}\right\|
\end{equation}

Limiting the condition number $\kappa(\mathbf{W})$ of the weight matrix of the network will lower the upper bound of the corresponding output response, which is aroused by input perturbations. Thus, reducing the condition number promotes the system's robustness to adversarial noise.

\section{Visualization of Per-Layer Pruning Ratio}

\begin{figure}[htbp]
	\centering
	\centering
	\begin{subfigure}{\linewidth}
		\centering
		\includegraphics[width=\linewidth]{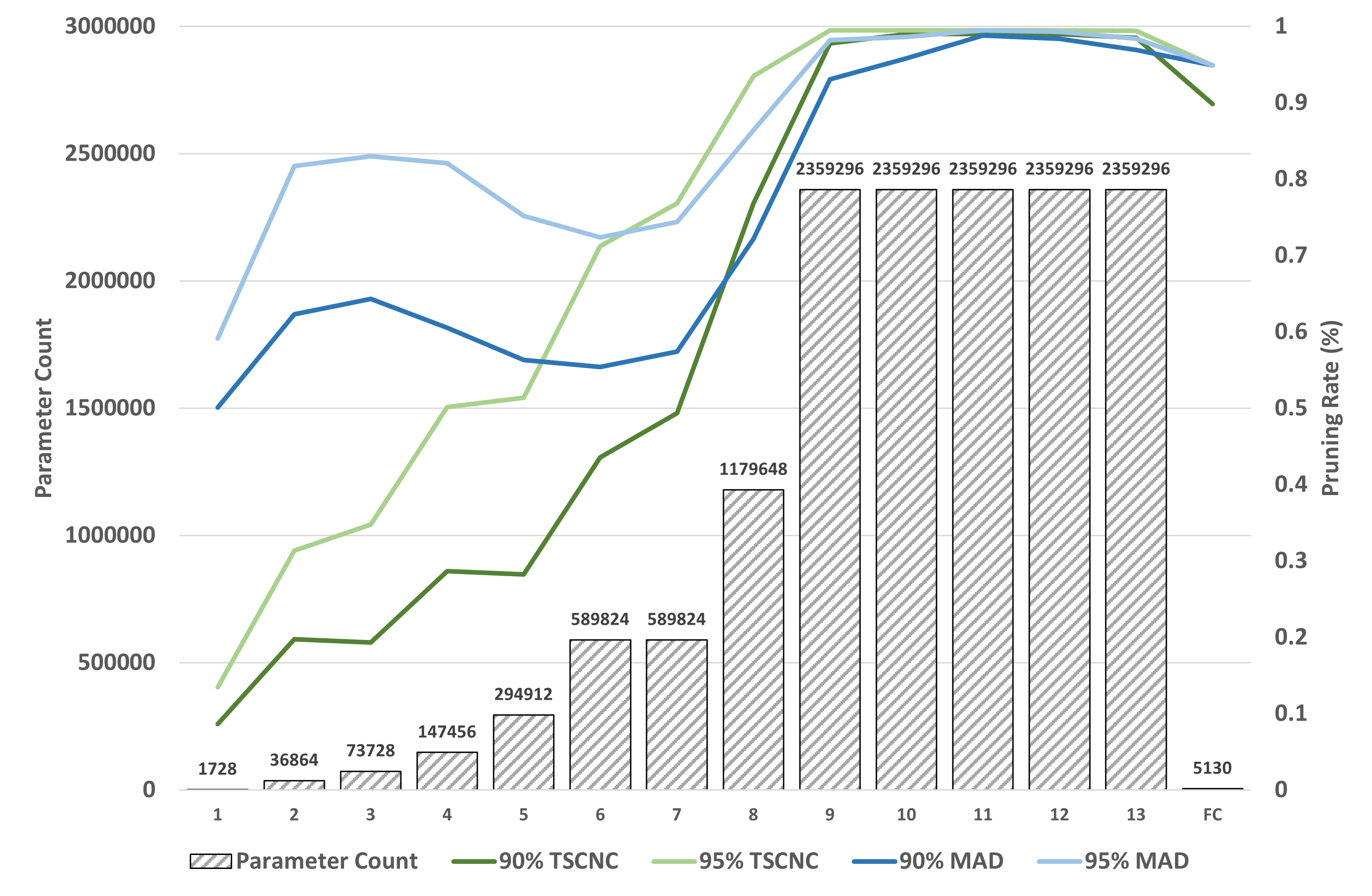}
		\caption{Layer-wise Pruning Ratio on VGG-16}
		\label{vgg16}
	\end{subfigure}
	\\
	\begin{subfigure}{\linewidth}
		\centering
		\includegraphics[width=\linewidth]{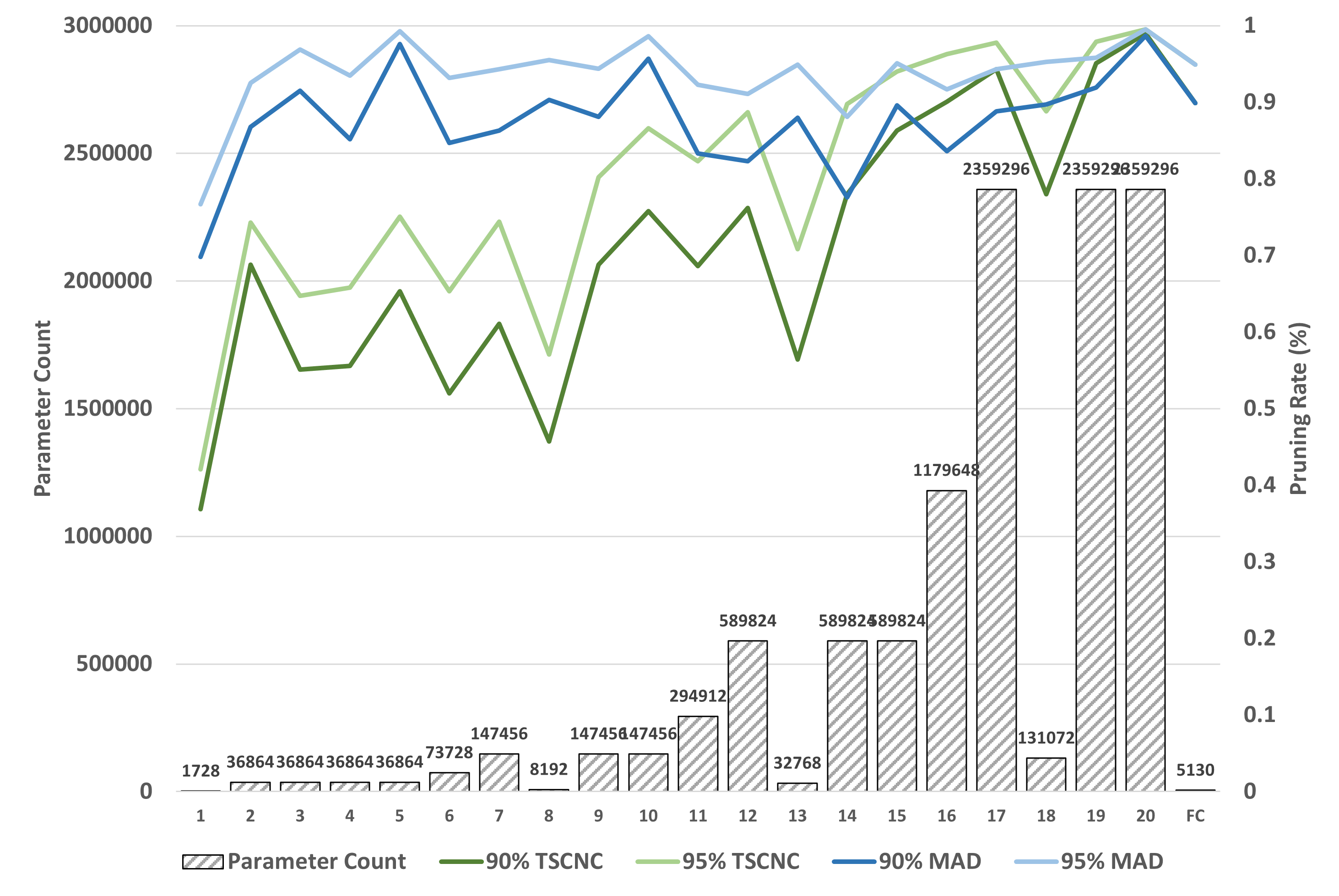}
		\caption{Layer-wise Pruning Ratio on ResNet-18}
		\label{rn18}
	\end{subfigure}
	\caption{Visualization of Layer-wise Pruning Ratio of TSCNC and MAD, under Sparsity of 90\% and 95\%, using (a) VGG-16 and (b) ResNet-18 on CIFAR-10.}
	\label{fig:prune_ratio}
\end{figure}

To compare how parameters are pruned, we visualize the per-layer pruning ratio of TSCNC and MAD in~\cref{fig:prune_ratio}.
As observed from figure, comparatively, TSCNC tends to prune less parameter in initial layers of the network, while it achieves higher pruning rate in layers with more number of parameters. As for MAD, it keeps pruning ratio above 50\% across all layers.

\section{More Details and Experiments of TSCNC}

\begin{table*}[htbp]
\centering
\small
\resizebox{0.8\textwidth}{!}{%
\begin{tabular}{c|l|cc|cc|cc|cc}
\toprule
 & \multicolumn{1}{c|}{} & \multicolumn{4}{c}{\textbf{CIFAR-10}} & \multicolumn{4}{c}{\textbf{CIFAR-100}} \\ \cmidrule(lr){1-2} \cmidrule(lr){3-6} \cmidrule(lr){7-10}
\multirow{2}{*}{Model} & \multicolumn{1}{c|}{Sparsity} & \multicolumn{2}{c}{90\%} & \multicolumn{2}{c}{95\%} & 
\multicolumn{2}{c}{90\%} & \multicolumn{2}{c}{95\%} \\ \cmidrule(lr){2-2} \cmidrule(lr){3-4} \cmidrule(lr){5-6} \cmidrule(lr){7-8} \cmidrule(lr){9-10} 
 & \multicolumn{1}{c|}{Method} & 
 MAD & TSCNC & 
 MAD & TSCNC & 
 MAD & TSCNC & 
 MAD & TSCNC \\ \midrule
\multirow{3}{*}{\shortstack{VGG-11 \\ \textit{(9.2M)}}} & 
Clean  & 68.07 & \textbf{79.65} & 65.01 & \textbf{77.98} & 
34.48 & \textbf{52.51} & 32.17 & \textbf{50.29} \\
 & FGSM& 46.38 & \textbf{53.52} & 44.09 & \textbf{52.39} & 
20.92 & \textbf{27.72} & 18.85 & \textbf{27.47} \\
 & PGD & 43.14 & \textbf{49.14} & 41.38 & \textbf{47.78} & 
19.34 & \textbf{25.66} & 17.77 & \textbf{24.97} \\
 \midrule
\multirow{3}{*}{\shortstack{ResNet-18 \\ \textit{(11.2M)}}} & 
 Clean& 
74.25 & \textbf{84.67} & 71.16 & \textbf{82.25} & 41.47 
& \textbf{54.93} & 34.77 & \textbf{51.08} \\
& FGSM& 
51.21 & \textbf{60.40} & 49.35 & \textbf{57.93} & 22.72 
& \textbf{31.26} & 19.20 & \textbf{28.88} \\
& PGD & 
47.84 & \textbf{54.07} & 46.49 & \textbf{53.70} & 21.52 
& \textbf{28.72} & 18.73 & \textbf{27.09} \\
\bottomrule
\end{tabular}%
}
\caption{Comparing the adversarial robustness accuracy on small-scaled datasets CIFAR-10 and CIFAR-100 with VGG-11 and ResNet-18. We control the sparsity of models to 90\% and 95\%, the bold indicates the best performance at the same sparsity under the same attack method.}
\label{tab:vgg11resnet18}
\end{table*}

\subsection{Results on VGG-11 and ResNet-18}

Performance of TSCNC and MAD on VGG-11 and ResNet-18 are presnted in~\cref{tab:vgg11resnet18}. Our method outperforms MAD by a large margin.

\subsection{Results on SVHN}

We've shown the gains achieved by {TSCNC} on CIFAR-10 dataset, CIFAR-100 dataset, and Tiny-Imagenet dataset. For SVHN dataset \cite{yuval2011reading}, this comprehensive advantage is even more obvious. As shown in table \ref{tab:svhn}, at a high pruning ratio, i.e., $95\%$ sparsity, the TSCNC outperformed HYDRA and MAD by $4.85\%$ and $4.72\%$ under AA attack on VGG.

\subsection{Comparison to ADMM and ATMC}

To validate the effectiveness of the proposed method, we compare the adversarial robustness on CIFAR-10 dataset with other baselines, i.e., ADMM\cite{ye2019adversarial} and ATMC \cite{gui2019model}. As shown in table \ref{tab:other}, It is worth noting that the TSCNC achieves an overwhelming advantage at a high pruning ratio. We obtain gains up to $8.7$, and $10.72$  percentage points in robust accuracy, compared to ADMM, and ATMC under AA attack method. This advantage may be traced to the joint constraint of the pruning, which prevents the parameter matrix from being ill-conditioned.
\begin{table*}[ht]
\centering
\begin{tabular}{c|l|ccc|ccc}
\toprule
 & \multicolumn{1}{c|}{} & \multicolumn{6}{c}{SVHN} \\ \cmidrule(lr){1-2} \cmidrule(lr){3-8}
\multirow{2}{*}{Model} & \multicolumn{1}{c|}{Sparsity} & \multicolumn{3}{c}{90\%} & \multicolumn{3}{c}{95\%} \\ \cmidrule(lr){2-2} \cmidrule(lr){3-5} \cmidrule(lr){6-8}
 & \multicolumn{1}{c|}{Method} & HYDRA & MAD & TSCNC  & HYDRA & MAD & TSCNC \\ \midrule
\multirow{5}{*}{\shortstack{VGG-16 \\ \textit{(14.7M)}}}
 & Clean& 87.90 & 92.80 & \textbf{93.05} & 85.50 & 91.90 & \textbf{92.91} \\
 & FGSM & 64.22 & 64.54 & \textbf{68.96} & 63.11 & 63.25 & \textbf{68.17} \\
 & PGD  & 54.61 & 55.46 & \textbf{58.89} & 52.12 & 53.44 & \textbf{57.70} \\
 & CW$_{\infty}$ & 49.72 & 49.85 & \textbf{52.17} & 48.21 & 48.73 & \textbf{51.69} \\
 & AA   & 47.55 & 48.87 & \textbf{51.79} & 46.03 & 46.16 & \textbf{50.88} \\ \midrule
\multirow{5}{*}{\shortstack{WRN-28-10 \\ \textit{(36.5M)}}}
 & Clean& 91.00 & 93.80 & \textbf{94.70} & --.-- & 92.87 & \textbf{93.32} \\
 & FGSM & 69.07 & 70.94 & \textbf{74.91} & 68.28 & 69.15 & \textbf{74.13} \\
 & PGD  & 60.34 & 60.28 & \textbf{62.46} & 59.07 & 58.75 & \textbf{61.89} \\
 & CW$_{\infty}$ & 56.42 & 56.71 & \textbf{58.43} & 54.34 & 54.87 & \textbf{57.30} \\
 & AA   & 54.45 & 54.57 & \textbf{56.55} & 52.16 & 52.02 & \textbf{55.94} \\ \bottomrule
\end{tabular}
\caption{Comparison of our approach with other pruning-based baseline methods.
We use SVHN dataset with VGG-16 and WideResNet networks, iterative
adversarial training from \cite{madry2017towards} for this experiment.}
\label{tab:svhn}
\end{table*}

\begin{table*}[]
\centering
\begin{tabular}{c|l|ccc|ccc}
\toprule
 & \multicolumn{1}{c|}{} & \multicolumn{6}{c}{CIFAR-10} \\ \cmidrule(lr){1-2} \cmidrule(lr){3-8}
\multirow{2}{*}{Model} & \multicolumn{1}{c|}{Sparsity} & \multicolumn{3}{c}{90\%}  & \multicolumn{3}{c}{95\%} \\ \cmidrule(lr){2-2} \cmidrule(lr){3-5} \cmidrule(lr){6-8} 
 & \multicolumn{1}{c|}{Method} & ADMM & ATMC & TSCNC   & ADMM & ATMC & TSCNC \\ \midrule
\multirow{4}{*}{\shortstack{VGG-16 \\ \textit{(14.7M)}}} & FGSM & 51.94 & 51.07 & \textbf{60.17}  & 48.45 & 49.37 & \textbf{58.74} \\
 & PGD & 44.89 & 44.81 & \textbf{53.27}  & 41.45 & 41.50 & \textbf{51.87} \\
 & SA & 46.71 & 46.61 & \textbf{55.80}  & 44.95 & 43.02 & \textbf{53.36} \\
 & AA & 39.27 & 40.46 & \textbf{49.49}  & 35.71 & 38.13 & \textbf{46.43} \\ \midrule
\multirow{4}{*}{\shortstack{WRN-28-10 \\ \textit{(36.5M)}}} & FGSM & 52.35 & 52.57 & \textbf{57.63}  & 50.81 & 50.93 & \textbf{56.32} \\
 & PGD & 48.77 & 50.15 & \textbf{54.91}  & 46.47 & 48.28 & \textbf{53.67} \\
 & CW$_{\infty}$ & 50.55 & 51.82 & \textbf{57.01}  & 47.96 & 48.39 & \textbf{55.97} \\
 & AA & 46.45 & 47.54 & \textbf{54.09}  & 42.17 & 44.03 & \textbf{53.54} \\ \bottomrule
\end{tabular}
\caption{Comparing the adversarial robustness accuracy on  CIFAR-10 with other baselines, ADMM and ATMC. We control the sparsity of models to $90\%$ and $95\%$, the bold expression denotes the best performance within the same sparsity under the same attack method.}
\label{tab:other}
\end{table*}

\end{document}